\documentclass{article}
\usepackage{ijcai26}

\usepackage{times}
\usepackage{soul}
\usepackage{url}
\usepackage[hidelinks]{hyperref}
\usepackage[utf8]{inputenc}
\usepackage[small]{caption}
\usepackage{graphicx}
\usepackage{amsmath}
\usepackage{amssymb}
\usepackage{amsthm}
\usepackage{booktabs}
\usepackage{multirow}
\usepackage{algorithm}
\usepackage{algorithmic}
\usepackage[table]{xcolor}
\usepackage[switch]{lineno}
\usepackage{tcolorbox}
\tcbuselibrary{listings,breakable}
\usepackage[ruled,vlined,linesnumbered,algo2e]{algorithm2e}
\usepackage{listings}
\definecolor{myblue}{RGB}{30,144,255}
\definecolor{mytitleindigo}{RGB}{75,0,130}
\definecolor{mytitlered}{RGB}{178,34,34}
\definecolor{mytitleorange}{RGB}{255,140,0}
\definecolor{mytitleyellow}{RGB}{218,165,32}
\definecolor{mytitlegreen}{RGB}{34,139,34}
\definecolor{mytitleblue}{RGB}{29,78,137}
\definecolor{mytitleviolet}{RGB}{148,0,211}

\title{Adaptive GoGI-Skip: Coupling Goal-Gradient Importance with Dynamic Uncertainty for Efficient Reasoning}

\author{
    Ren Zhuang
    \affiliations
    School of Information Science and Technology, Hangzhou Normal University
    \emails
    venomrko97@gmail.com, 2022112011050@stu.hznu.edu.cn
}

\begin{document}

\maketitle


\begin{abstract}
Chain-of-Thought (CoT) prompting trades inference speed for reasoning accuracy. Existing compressors force a compromise as static gradient techniques treat tokens independently, severing sequential logic, while uncertainty-based pruning ignores the final answer. We introduce Adaptive GoGI-Skip, a framework that resolves this tension by non-linearly coupling Goal-Gradient Importance (GoGI) with Adaptive Dynamic Skipping (ADS). GoGI quantifies each token's functional contribution to answer correctness via gradient sensitivity. ADS leverages runtime entropy to dynamically modulate the GoGI threshold, preserving low-gradient tokens essential for structural coherence at high-uncertainty junctions. Trained on 7,472 MATH traces, our policy transfers zero-shot to AIME, GPQA, and GSM8K, reducing token volume by $>$45\% and accelerating inference up to 2.0$\times$ without accuracy loss. These results suggest that thinking-optimal compression demands synergy between teleological goals and epistemic uncertainty.
\end{abstract}
\section{Introduction}
\label{sec:introduction}

\begin{figure}[t]
  \centering
  \includegraphics[width=\columnwidth]{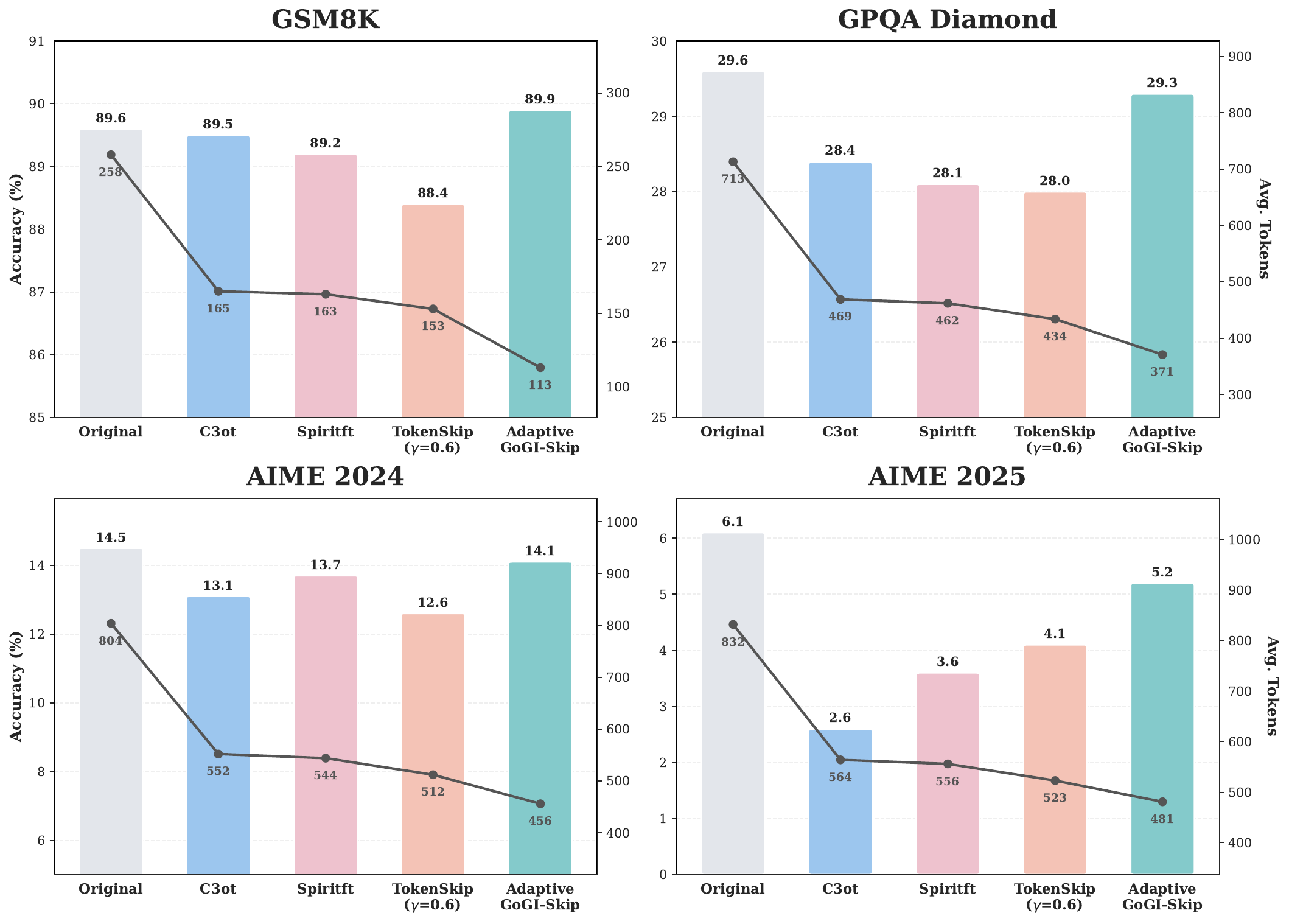}
  \caption{\textbf{Accuracy--Token Trade-off on Gemma3-4B-Instruct.} Adaptive GoGI-Skip shifts the trade-off curve, achieving superior accuracy retention compared to static baselines. The method dominates on rigorous benchmarks, preserving reasoning capability despite 45\% token reduction.}
  \label{fig:teaser_viz}
\end{figure}

Chain-of-Thought (CoT) prompting unlocks advanced reasoning in Large Language Models~\cite{wei2022chain} but incurs prohibitive inference latency and memory costs~\cite{qu2025survey,sui2025stop}. Standard efficiency paradigms~\cite{shazeer2017outrageously,katharopoulos2020transformers,su2024dualformer,wang2024make} overlook CoT's unique structural redundancy.

CoT traces often contain repetition, circular validation, and excessive elaboration~\cite{fatemi2025concise,sui2025stop}, sometimes degrading accuracy on simpler tasks~\cite{yang2025towards}. A thinking-optimal equilibrium, where models reason only as much as strictly necessary, requires a compression paradigm that is both goal-aware and dynamically adaptive.

Existing Supervised Fine-Tuning (SFT) compression approaches~\cite{xia2025tokenskip} rely on generic importance metrics like semantic similarity~\cite{pan2024llmlingua} or perplexity~\cite{cui2025stepwise} that often mask structurally critical yet semantically simple tokens. Static compression rates compound the problem by ignoring heterogeneous step difficulty~\cite{ge2024model}.

A straightforward combination of gradient-based attribution and adaptive computation can be brittle. In long-context generation, gradient signals are often sparse and noisy~\cite{wang2024gradient}, and pruning solely by gradient magnitude may disrupt the sequential dependencies required for multi-step deduction. Conversely, uncertainty-based criteria capture local ambiguity but do not necessarily reflect a token's contribution to final answer correctness. These observations motivate coupling the two signals: using epistemic uncertainty to dynamically regulate the teleological goal-alignment threshold.

\begin{figure*}[t]
  \centering
  \includegraphics[width=\textwidth]{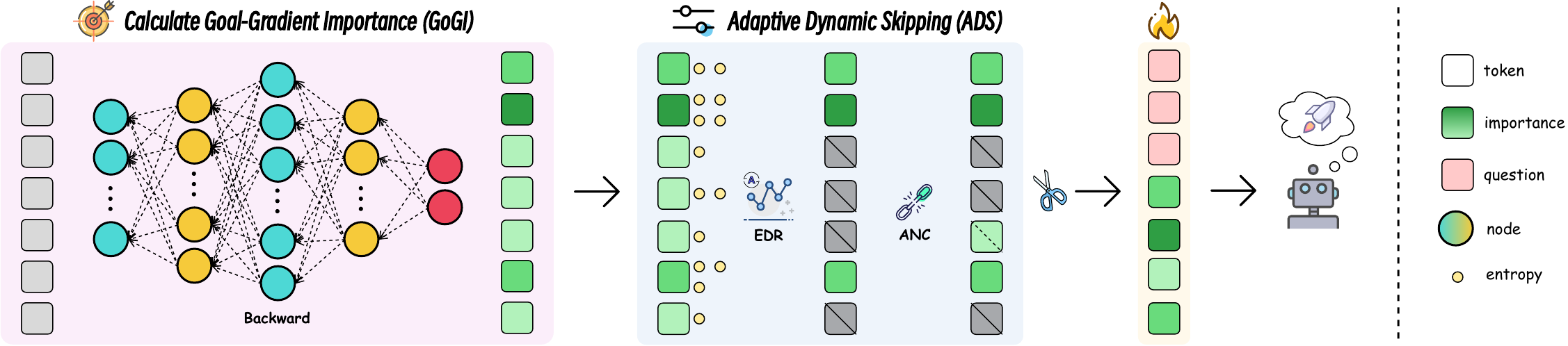}
  \caption{\textbf{The Adaptive GoGI-Skip Framework.} (a) \textbf{GoGI} measures token utility via $\nabla \mathcal{L}_{\text{ans}}$. (b) \textbf{ADS} combines GoGI with runtime uncertainty and local stability to decide $K_t$. (c) \textbf{SFT} distills the policy into model weights for overhead-free speedups.}
  \label{fig:framework}
\end{figure*}

To bridge this gap, we propose \textbf{Adaptive GoGI-Skip}, a framework for learning efficient, adaptive CoT compression. As Figure~\ref{fig:framework} illustrates, our approach integrates two synergistic innovations. Goal-Gradient Importance (GoGI) is a metric ($\mathcal{G}(x_i, l) = || \nabla_{\mathbf{h}_i^l} \mathcal{L}_{\text{ans}} ||_1$) that quantifies each token's functional utility via its gradient influence on the final answer loss. Adaptive Dynamic Skipping (ADS) employs Entropy-Driven Rate (EDR) Regulation, using predictive entropy $H_t$ to adjust retention rates, and an Adaptive N-Constraint (ANC) that modulates consecutive deletions based on local complexity, preserving coherence.

Our key contributions are summarized below:
\begin{itemize}
  \item We introduce Adaptive GoGI-Skip, a goal-aligned and uncertainty-regulated CoT compression framework that couples Goal-Gradient Importance (GoGI) with Adaptive Dynamic Skipping (ADS) to remove functionally redundant tokens while preserving reasoning-critical structure.
  \item We establish zero-shot universality by training a single compression policy on 7,472 MATH traces and transferring it across diverse benchmarks (AIME, GPQA, GSM8K) and model families (Gemma, Qwen) without task-specific tuning.
  \item We push the accuracy--efficiency frontier, achieving $>$45\% token reduction and up to 2.0$\times$ speedup while matching or improving original accuracy on rigorous evaluations (Figure~\ref{fig:teaser_viz}).
\end{itemize}
\section{Related Work}
\label{sec:related_work}

CoT reasoning's efficacy incurs substantial computational cost~\cite{qu2025survey,sui2025stop}. Optimization efforts bifurcate into SFT-based compression and alternative reasoning paradigms.

\paragraph{Token Importance Estimation.}
Existing compression methods rely on proxy metrics for token utility. External distillers such as C3OT~\cite{kang2025c3ot} leverage GPT-4. Internal methods instead target semantic redundancy, and TokenSkip~\cite{xia2025tokenskip} builds on LLMLingua~\cite{pan2024llmlingua}. Other approaches exploit activation sparsity, including LazyLLM~\cite{fu2025lazyllm} and SlimInfer~\cite{long2025sliminfer}. SPIRIT~\cite{cui2025stepwise} uses perplexity-based signals. These heuristics decouple importance from functional relevance and can remove structurally vital yet semantically simple logical connectors~\cite{razzhigaev2025llm}. Gradient-based saliency is well-established in interpretability~\cite{simonyan2013deep,sundararajan2017axiomatic}, but naive application to pruning fails since attribution does not necessarily reflect generation stability. High-gradient tokens are often high-entropy surprises that, if pruned, cause catastrophic drift; discarding low-gradient tokens can sever reasoning's connective tissue.

GoGI departs from generic attribution by non-linearly coupling gradient magnitude with predictive entropy, creating a dynamic threshold that adapts to the chain's structural fragility.

\paragraph{Adaptive Compression Strategies.}
Current SFT-based approaches employ static compression rates~\cite{xia2025tokenskip} or coarse step-level skipping~\cite{wang2025r1compress}. These rigid policies struggle to accommodate heterogeneous reasoning complexity. Recent work explores adaptive reasoning steps~\cite{wang2025adareasoner} or compute allocation~\cite{manvi2024adaptive}, but often requires complex routers.

Adaptive GoGI-Skip addresses this via ADS. By modulating retention rates based on real-time predictive entropy and regulating consecutive deletions via local context, ADS achieves fine-grained balance between sparsity and coherence unattainable by static baselines.

\paragraph{Alternative Efficiency Paradigms.}
Parallel tracks explore different trade-offs. Latent-space reasoning~\cite{saunshi2025reasoning,hao2024training} bypasses linguistic decoding entirely but sacrifices interpretability~\cite{ma2025reasoning}. Reinforcement Learning (RL) approaches~\cite{luo2025o1,shen2025dast} integrate length penalties into reward functions, yet RL often fails to incentivize genuine reasoning beyond the base model~\cite{yue2025does}, induces training instability~\cite{melo2025stabilizing}, and is prone to reward hacking~\cite{gao2024designing} or increased hallucination~\cite{yao2025reasoning}.

Our SFT-based approach offers a distinct advantage through stable, direct optimization that learns precise compression policies without RL's fragility. It also serves as a complementary post-processor that trims verbose traces produced by RL-optimized reasoners.
\section{Adaptive GoGI-Skip for Reasoning-Intensive CoT Optimization}
\label{sec:methodology}

Adaptive GoGI-Skip is an SFT framework for dynamic pruning toward Reasoning-Intensive CoT Optimization (RICO). The method combines GoGI, which measures each token's contribution to the answer loss, with ADS, which adapts the retention policy from predictive uncertainty and local stability. This coupling yields finer-grained compression than prior approaches.

\paragraph{Notation.} We denote sequences by $(x_t)_{t=1}^m$ and write vectors in bold lowercase. The hidden state of token $t$ at layer $l$ is $\mathbf{h}_t^l$. Key metrics include the GoGI score $\mathcal{G}_t$, predictive entropy $H_t$, dynamic retention rate $\gamma_t$, dynamic GoGI threshold $\tau_t$, and the adaptive N-constraint $N_t$. The keep decision is $K_t \in \{0,1\}$. We denote the answer loss by $\mathcal{L}_{\text{ans}}$ and model parameters by $\theta$. The operator $||\cdot||_1$ denotes the L1 norm, $\nabla$ the gradient, and $Q_p$ the $p$-th percentile operator. We use $[\cdot]_{a}^{b}$ for clipping to $[a,b]$, $\lfloor\cdot\rfloor$ for the floor, and $\mathbb{I}$ for the indicator function. We write $\lor$ for logical OR and $\land$ for logical AND.

\subsection{Goal-Gradient Importance (GoGI)}
\label{ssec:gogi}

Existing importance metrics~\cite{xia2025tokenskip,pan2024llmlingua,cui2025stepwise} ignore the reasoning objective. GoGI directly measures each token's contribution to the target objective via gradient-based sensitivity analysis, grounded in the principle that gradients inherently quantify functional importance.

Given a CoT sequence $c = (x_t)_{t=1}^m$ and answer $A = (a_t)_{t=1}^k$, the answer prediction loss $\mathcal{L}_{\text{ans}}$ for model $M_\theta$ is:
\begin{equation}
    \mathcal{L}_{\text{ans}}(A | c; \theta) = - \sum_{j=1}^{k} \log P_\theta(a_j | c, a_{<j})
    \label{eq:ans_loss}
\end{equation}
The GoGI score for token $x_t$ at target layer $l^*$ is the L1 norm of the gradient:
\begin{equation}
    \label{eq:gogi}
    \mathcal{G}_t^{(l^*)} \triangleq \left\| \frac{\partial \mathcal{L}_{\text{ans}}(A | c; \theta)}{\partial \mathbf{h}_t^{l^*}} \right\|_1
\end{equation}
This score measures the local sensitivity of the final answer loss to the intermediate representation $\mathbf{h}_t^{l^*}$.

\begin{figure}[h]
    \centering
    \includegraphics[width=\linewidth]{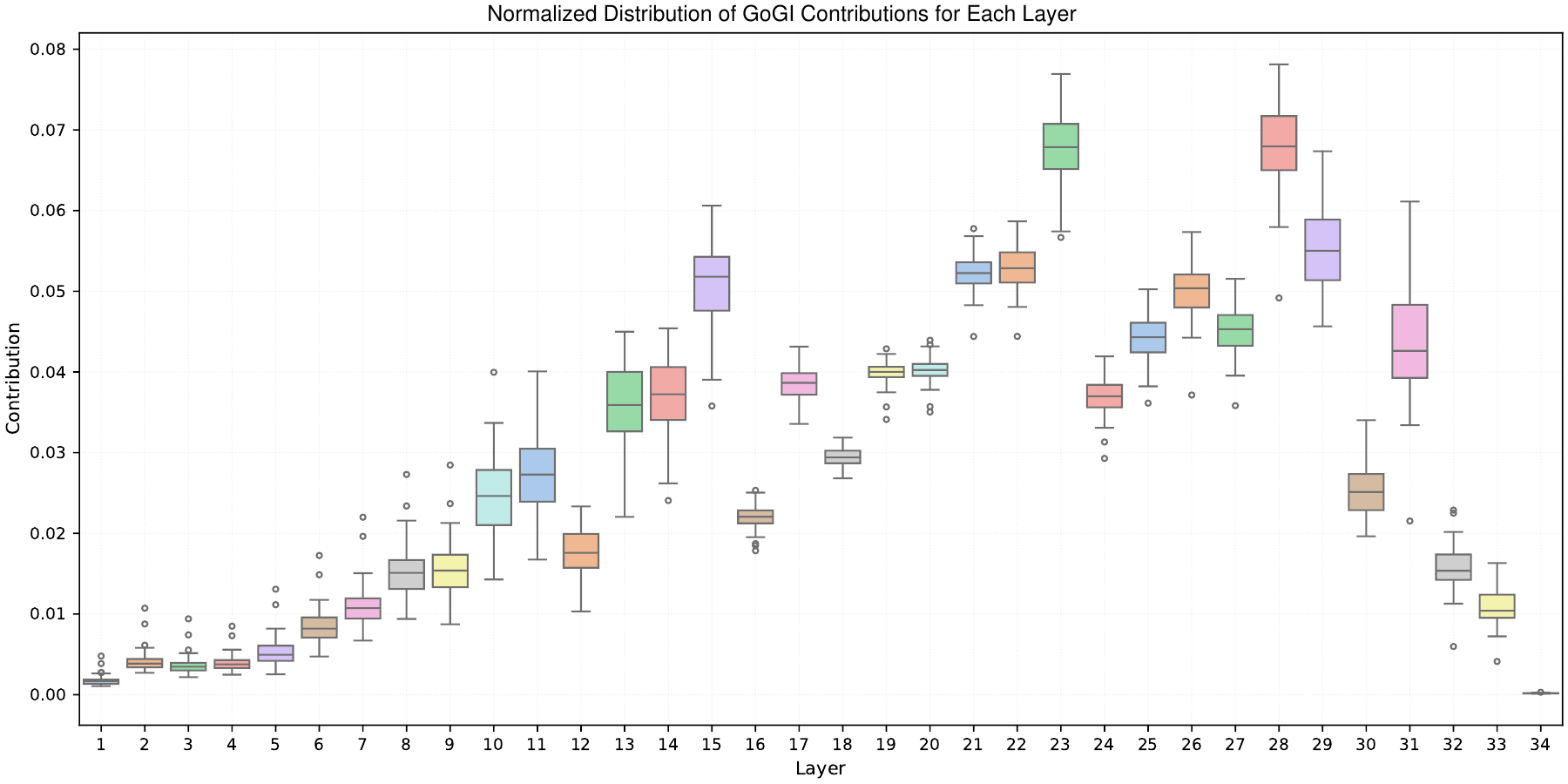}
    \caption{\textbf{Layer-wise normalized gradients on Gemma3-4B-Instruct.} Layers 23 and 28 show similar peak magnitudes.}
    \label{fig:layer_contribution_variability}
\end{figure}

We select the target layer $l^*$ by analyzing layer-wise gradient contributions. Peak sensitivity occurs in mid-to-late layers. Within this high-contribution region, downstream performance is empirically insensitive to the exact choice of $l^*$. On Gemma3-4B-Instruct, for example, Layers 23 and 28 exhibit comparable peaks (Figure~\ref{fig:layer_contribution_variability}). We therefore select Layer 23 as it balances the local and global signals for GoGI computation; analogous analyses for other models are reported in the Supplementary Material.

\begin{algorithm}[t]
    \caption{Adaptive GoGI-Skip Pruning}
    \label{alg:gogi_skip}
    \begin{algorithmic}[1]
        \renewcommand{\algorithmicrequire}{\textbf{Input:}}
        \renewcommand{\algorithmicensure}{\textbf{Output:}}
        \REQUIRE CoT sequence $c = (x_t)_{t=1}^m$, answer $A$, model $M_\theta$
        \ENSURE Compressed sequence $c'_{\text{comp}}$, keep mask $(K_t)_{t=1}^m$
        \STATE Compute $\mathcal{L}_{\text{ans}}(A|c;\theta)$ via forward pass
        \STATE $\mathcal{G}_t \gets \|\nabla_{\mathbf{h}_t^{l^*}} \mathcal{L}_{\text{ans}}\|_1$ for all $t$ \COMMENT{GoGI scores}
        \STATE $H_t \gets -\sum_v p_{t,v} \log p_{t,v}$ for all $t$ \COMMENT{Entropy}
        \STATE Determine $\mathcal{I}_{\text{valid}}$ by excluding non-substantive tokens
        \STATE $C_0 \gets 0$ \COMMENT{Consecutive prune counter}
        \FOR{$t = 1$ to $m$}
        \IF{$t \notin \mathcal{I}_{\text{valid}}$}
        \STATE $K_t \gets 1$
        \STATE $C_t \gets 0$
        \STATE \textbf{continue}
        \ENDIF
        \STATE $\hat{H}_t \gets \mathcal{M}(H_t; \theta_{\mathcal{M}})$
        \STATE $\gamma_t \gets [\gamma_{\min} + (\gamma_{\max} - \gamma_{\min}) \cdot \hat{H}_t]_{\gamma^{\text{abs}}_{\text{min}}}^{\gamma^{\text{abs}}_{\text{max}}}$ \COMMENT{EDR}
        \STATE $\tau_t \gets Q_{(1-\gamma_t)\times 100}(\{\mathcal{G}_j \mid j \in \mathcal{I}_{\text{valid}}\})$
        \STATE $\bar{H}_{t, W} \gets \frac{1}{W}\sum_{j=\max(1,t-\lfloor W/2\rfloor)}^{\min(m,t+\lfloor W/2\rfloor)} H_j$
        \STATE $\hat{\bar{H}}_{t, W} \gets \mathcal{M}'(\bar{H}_{t, W}; \theta_{\mathcal{M}'})$
        \STATE $\tilde{N}_t \gets N_{\min} + (N_{\max} - N_{\min})(1 - \hat{\bar{H}}_{t, W})$
        \STATE $N_t \gets \lfloor [\tilde{N}_t]_{N_{\min}}^{N_{\max}} + 0.5 \rfloor$ \COMMENT{ANC}
        \STATE $\mathcal{G}'_t \gets f_{\text{weight}}(\mathcal{G}_t, x_t)$ \COMMENT{Optional type-weighting}
        \IF{$\mathcal{G}'_t \ge \tau_t$ \OR $C_{t-1} + 1 \ge N_t$}
        \STATE $K_t \gets 1$
        \STATE $C_t \gets 0$
        \ELSE
        \STATE $K_t \gets 0$
        \STATE $C_t \gets C_{t-1} + 1$
        \ENDIF
        \ENDFOR
        \STATE $c'_{\text{comp}} \gets (x_t)_{K_t = 1}$
        \RETURN $c'_{\text{comp}}$, $(K_t)_{t=1}^m$
    \end{algorithmic}
\end{algorithm}

\begin{table*}[t]
    \centering
    \begin{tabular}{l|cccc|cccc}
        \toprule
        \multirow{2}{*}{\raisebox{-0.6ex}{Method}}                           & \multicolumn{4}{c|}{AIME 2025~\cite{AIME25}}          & \multicolumn{4}{c}{AIME 2024~\cite{AIME24}}                                                                                                                                                                        \\
        \cmidrule(lr){2-5}\cmidrule(lr){6-9}
                                                          & Acc.$\uparrow$                                        & Ratio $\downarrow$                              & Tokens$\downarrow$ & Speedup$\uparrow$    & Acc.$\uparrow$                                      & Ratio $\downarrow$ & Tokens$\downarrow$ & Speedup$\uparrow$    \\
        \midrule
        Original                                          & 6.1$_{{(0.0\downarrow)}}$                             & 1.0                                             & 832                & 1.0$\times$          & 14.5$_{{(0.0\downarrow)}}$                          & 1.0                & 804                & 1.0$\times$          \\
        Prompting                                         & 1.8$_{{(4.3\downarrow)}}$                             & 0.9                                             & 764                & 1.0$\times$          & 10.2$_{{(4.3\downarrow)}}$                          & 0.9                & 728                & 1.1$\times$          \\
        C3ot                                              & 2.6$_{{(3.5\downarrow)}}$                             & 0.7                                             & 564                & 1.4$\times$          & 13.1$_{{(1.4\downarrow)}}$                          & 0.7                & 552                & 1.4$\times$          \\
        Spiritft                                          & 3.6$_{{(2.5\downarrow)}}$                             & 0.6                                             & 556                & 1.5$\times$          & 13.7$_{{(0.8\downarrow)}}$                          & 0.7                & 544                & 1.4$\times$          \\
        TokenSkip ($\gamma=0.6$)                          & 4.1$_{{(2.0\downarrow)}}$                             & 0.6                                             & 523                & 1.6$\times$          & 12.6$_{{(1.9\downarrow)}}$                          & 0.6                & 512                & 1.5$\times$          \\
        TokenSkip ($\gamma=0.5$)                          & 2.7$_{{(3.4\downarrow)}}$                             & 0.6                                             & 457                & 1.7$\times$          & 11.3$_{{(3.2\downarrow)}}$                          & 0.5                & 448                & 1.6$\times$          \\
        \rowcolor[HTML]{E6E6E6} Adaptive GoGI-Skip (Ours) & \textbf{5.2$_{\textbf{(0.9}\downarrow\textbf{)}}$}    & \textbf{0.6}                                    & \textbf{481}       & \textbf{1.6$\times$} & \textbf{14.1$_{\textbf{(0.4}\downarrow\textbf{)}}$} & \textbf{0.6}       & \textbf{456}       & \textbf{1.7$\times$} \\
        \midrule
        \midrule
        \multirow{2}{*}{\raisebox{-0.6ex}{Method}}                           & \multicolumn{4}{c|}{GPQA Diamond~\cite{rein2024gpqa}} & \multicolumn{4}{c}{GSM8K~\cite{cobbe2021gsm8k}}                                                                                                                                                                    \\
        \cmidrule(lr){2-5}\cmidrule(lr){6-9}
                                                          & Acc.$\uparrow$                                        & Ratio$\downarrow$                               & Tokens$\downarrow$ & Speedup$\uparrow$    & Acc.$\uparrow$                                      & Ratio$\downarrow$  & Tokens$\downarrow$ & Speedup$\uparrow$    \\
        \midrule
        Original                                          & 29.6$_{{(0.0\downarrow)}}$                            & 1.0                                             & 713                & 1.0$\times$          & 89.6$_{{(0.0\downarrow)}}$                          & 1.0                & 258                & 1.0$\times$          \\
        Prompting                                         & 21.7$_{{(7.9\downarrow)}}$                            & 0.9                                             & 651                & 1.0$\times$          & 80.2$_{{(9.4\downarrow)}}$                          & 0.9                & 225                & 1.1$\times$          \\
        C3ot                                              & 28.4$_{{(1.2\downarrow)}}$                            & 0.7                                             & 469                & 1.5$\times$          & 89.5$_{{(0.1\downarrow)}}$                          & 0.6                & 165                & 1.5$\times$          \\
        Spiritft                                          & 28.1$_{{(1.5\downarrow)}}$                            & 0.6                                             & 462                & 1.5$\times$          & 89.2$_{{(0.4\downarrow)}}$                          & 0.6                & 163                & 1.5$\times$          \\
        TokenSkip ($\gamma=0.6$)                          & 28.0$_{{(1.6\downarrow)}}$                            & 0.6                                             & 434                & 1.6$\times$          & 88.4$_{{(1.2\downarrow)}}$                          & 0.6                & 153                & 1.6$\times$          \\
        TokenSkip ($\gamma=0.5$)                          & 26.5$_{{(3.1\downarrow)}}$                            & 0.5                                             & 378                & 1.8$\times$          & 84.3$_{{(5.0\downarrow)}}$                          & 0.5                & 133                & 1.9$\times$          \\
        \rowcolor[HTML]{E6E6E6} Adaptive GoGI-Skip (Ours) & \textbf{29.3$_{\textbf{(0.3}\downarrow\textbf{)}}$}   & \textbf{0.5}                                    & \textbf{371}       & \textbf{1.8$\times$} & \textbf{89.9$_{\textbf{(0.3}\uparrow\textbf{)}}$}   & \textbf{0.4}       & \textbf{113}       & \textbf{1.9$\times$} \\
        \bottomrule
    \end{tabular}
    \caption{\textbf{Main Efficiency Results on Gemma3-4B-Instruct.} Adaptive GoGI-Skip achieves consistent speedups (1.6--1.9$\times$) with negligible accuracy loss ($<$0.9\%), outperforming static baselines that degrade significantly on reasoning-intensive tasks.}
    \label{tab:main_results_gemma4b}
\end{table*}

\begin{table*}[t]
    \centering
    \resizebox{\textwidth}{!}{
        \begin{tabular}{l|cccccccc|cccccccc}
            \toprule
                                                       & \multicolumn{8}{c|}{Gemma3-1B-It}          & \multicolumn{8}{c}{Gemma3-12B-It}                                                                                                                                                                                                                                                                                                                                                                              \\
            \cmidrule(lr){2-9}\cmidrule(lr){10-17}
            \multirow{2}{*}{\raisebox{+1.6ex}{Method}}                    & \multicolumn{2}{c}{AIME\textquotesingle25} & \multicolumn{2}{c}{AIME\textquotesingle24} & \multicolumn{2}{c}{GPQA} & \multicolumn{2}{c|}{GSM8K} & \multicolumn{2}{c}{AIME\textquotesingle25} & \multicolumn{2}{c}{AIME\textquotesingle24} & \multicolumn{2}{c}{GPQA} & \multicolumn{2}{c}{GSM8K}                                                                                                                                                            \\
            \cmidrule(lr){2-3}\cmidrule(lr){4-5}\cmidrule(lr){6-7}\cmidrule(lr){8-9}\cmidrule(lr){10-11}\cmidrule(lr){12-13}\cmidrule(lr){14-15}\cmidrule(lr){15-17}
                                                       & Acc.                                       & SU.                                        & Acc.                     & SU.                        & Acc.                                       & SU.                                        & Acc.                     & SU.                       & Acc.         & SU.                  & Acc.          & SU.                  & Acc.          & SU.                  & Acc.          & SU.                  \\
            \midrule
            Original                                   & 0.7                                        & 1.0$\times$                                & 3.1                      & 1.0$\times$                & 18.7                                       & 1.0$\times$                                & 70.2                     & 1.0$\times$               & 8.1          & 1.0$\times$          & 22.7          & 1.0$\times$          & 38.4          & 1.0$\times$          & 95.2          & 1.0$\times$          \\
            Prompting                                  & 0.1                                        & 1.1$\times$                                & 1.2                      & 1.1$\times$                & 12.5                                       & 1.1$\times$                                & 65.7                     & 1.1$\times$               & 4.2          & 1.1$\times$          & 15.3          & 1.1$\times$          & 28.7          & 1.1$\times$          & 85.6          & 1.1$\times$          \\
            C3ot                                       & 0.5                                        & 1.4$\times$                                & 2.3                      & 1.4$\times$                & 16.1                                       & 1.4$\times$                                & 68.7                     & 1.5$\times$               & 7.4          & 1.4$\times$          & 21.6          & 1.5$\times$          & 37.4          & 1.5$\times$          & 94.1          & 1.5$\times$          \\
            Spiritft                                   & 0.4                                        & 1.4$\times$                                & 2.1                      & 1.4$\times$                & 17.3                                       & 1.5$\times$                                & 69.6                     & 1.5$\times$               & 7.2          & 1.5$\times$          & 21.2          & 1.5$\times$          & 36.6          & 1.5$\times$          & 93.8          & 1.6$\times$          \\
            TokenSkip ($\gamma$=0.6)                   & 0.5                                        & 1.5$\times$                                & 2.2                      & 1.5$\times$                & 16.5                                       & 1.5$\times$                                & 69.9                     & 1.6$\times$               & 7.5          & 1.5$\times$          & 21.9          & 1.6$\times$          & 37.1          & 1.6$\times$          & 94.3          & 1.6$\times$          \\
            TokenSkip ($\gamma$=0.5)                   & 0.3                                        & 1.6$\times$                                & 1.9                      & 1.6$\times$                & 15.8                                       & 1.6$\times$                                & 64.6                     & 1.7$\times$               & 6.8          & 1.6$\times$          & 21.2          & 1.8$\times$          & 36.1          & 1.7$\times$          & 92.7          & 1.8$\times$          \\
            \rowcolor[HTML]{E6E6E6} Adaptive GoGI-Skip & \textbf{0.6}                               & \textbf{1.5$\times$}                       & \textbf{2.9}             & \textbf{1.5$\times$}       & \textbf{18.4}                              & \textbf{1.6$\times$}                       & \textbf{70.2}            & \textbf{1.8$\times$}      & \textbf{8.0} & \textbf{1.6$\times$} & \textbf{22.9} & \textbf{1.7$\times$} & \textbf{38.4} & \textbf{1.7$\times$} & \textbf{95.5} & \textbf{2.0$\times$} \\
            \midrule
            \midrule
                                                       & \multicolumn{8}{c|}{Qwen2.5-3B-It}         & \multicolumn{8}{c}{Qwen2.5-7B-It}                                                                                                                                                                                                                                                                                                                                                                              \\
            \cmidrule(lr){2-9}\cmidrule(lr){10-17}
            \multirow{2}{*}{\raisebox{+1.6ex}{Method}}                    & \multicolumn{2}{c}{AIME\textquotesingle25} & \multicolumn{2}{c}{AIME\textquotesingle24} & \multicolumn{2}{c}{GPQA} & \multicolumn{2}{c|}{GSM8K} & \multicolumn{2}{c}{AIME\textquotesingle25} & \multicolumn{2}{c}{AIME\textquotesingle24} & \multicolumn{2}{c}{GPQA} & \multicolumn{2}{c}{GSM8K}                                                                                                                                                            \\
            \cmidrule(lr){2-3}\cmidrule(lr){4-5}\cmidrule(lr){6-7}\cmidrule(lr){8-9}\cmidrule(lr){10-11}\cmidrule(lr){12-13}\cmidrule(lr){14-15}\cmidrule(lr){15-17}
                                                       & Acc.                                       & SU.                                        & Acc.                     & SU.                        & Acc.                                       & SU.                                        & Acc.                     & SU.                       & Acc.         & SU.                  & Acc.          & SU.                  & Acc.          & SU.                  & Acc.          & SU.                  \\
            \midrule
            Original                                   & 2.1                                        & 1.0$\times$                                & 7.5                      & 1.0$\times$                & 25.3                                       & 1.0$\times$                                & 78.5                     & 1.0$\times$               & 6.2          & 1.0$\times$          & 18.1          & 1.0$\times$          & 35.7          & 1.0$\times$          & 84.5          & 1.0$\times$          \\
            Prompting                                  & 0.8                                        & 1.1$\times$                                & 4.1                      & 1.1$\times$                & 17.2                                       & 1.1$\times$                                & 65.3                     & 1.1$\times$               & 2.8          & 1.1$\times$          & 11.2          & 1.1$\times$          & 25.8          & 1.1$\times$          & 77.1          & 1.1$\times$          \\
            C3ot                                       & 1.7                                        & 1.4$\times$                                & 6.8                      & 1.4$\times$                & 23.5                                       & 1.5$\times$                                & 76.1                     & 1.5$\times$               & 5.6          & 1.5$\times$          & 17.0          & 1.5$\times$          & 34.3          & 1.5$\times$          & 82.7          & 1.5$\times$          \\
            Spiritft                                   & 1.6                                        & 1.5$\times$                                & 6.5                      & 1.5$\times$                & 23.0                                       & 1.5$\times$                                & 75.2                     & 1.5$\times$               & 5.3          & 1.5$\times$          & 16.6          & 1.5$\times$          & 33.6          & 1.5$\times$          & 88.2          & 1.6$\times$          \\
            TokenSkip ($\gamma$=0.6)                   & 1.8                                        & 1.5$\times$                                & 6.9                      & 1.6$\times$                & 23.4                                       & 1.6$\times$                                & 76.5                     & 1.6$\times$               & 5.7          & 1.6$\times$          & 17.2          & 1.6$\times$          & 34.1          & 1.6$\times$          & 83.1          & 1.6$\times$          \\
            TokenSkip ($\gamma$=0.5)                   & 1.4                                        & 1.6$\times$                                & 5.8                      & 1.6$\times$                & 22.5                                       & 1.7$\times$                                & 73.8                     & 1.9$\times$               & 5.4          & 1.7$\times$          & 15.8          & 1.7$\times$          & 32.9          & 1.7$\times$          & 81.5          & 1.9$\times$          \\
            \rowcolor[HTML]{E6E6E6} Adaptive GoGI-Skip & \textbf{2.0}                               & \textbf{1.5$\times$}                       & \textbf{7.3}             & \textbf{1.5$\times$}       & \textbf{24.6}                              & \textbf{1.7$\times$}                       & \textbf{78.9}            & \textbf{1.8$\times$}      & \textbf{6.1} & \textbf{1.5$\times$} & \textbf{18.1} & \textbf{1.5$\times$} & \textbf{35.7} & \textbf{1.6$\times$} & \textbf{85.1} & \textbf{1.9$\times$} \\
            \bottomrule
        \end{tabular}
    }
    \caption{\textbf{Zero-shot generalization across models and scales.} Adaptive GoGI-Skip generalizes across Gemma and Qwen models from 1B to 12B. Efficiency gains increase with model size, consistent with stronger separation between reasoning signal and noise at larger scales.}
    \label{tab:generalization_models}
\end{table*}

\subsection{Adaptive Dynamic Skipping (ADS)}
\label{ssec:ads}

ADS dynamically adjusts the pruning strategy based on runtime context via EDR and ANC components.

\subsubsection{Entropy-Driven Rate (EDR) Regulation}
\label{sssec:edr}

EDR regulation leverages predictive entropy $H_t$ (Eq.~\ref{eq:entropy}) as an information density signal. Since only a fraction of CoT tokens exhibit high entropy---signaling critical branching points~\cite{wang2025beyond}---high $H_t$ warrants conservative pruning. Crucially, $H_t$ and $\mathcal{G}_t$ provide complementary signals.

\begin{equation}
    H_t = -\sum_{v \in \mathcal{V}} p_{t,v} \log p_{t,v},
    \label{eq:entropy}
\end{equation}
where $p_{t,v} = P_\theta(X_{t+1}=v | x_1, \ldots, x_t)$.
We map $H_t$ to $\hat{H}_t \in [0, 1]$ via $\mathcal{M}(\cdot; \theta_{\mathcal{M}})$ (see Supplementary Material):
\begin{equation}
    \hat{H}_t = \mathcal{M}(H_t; \theta_{\mathcal{M}})
\end{equation}
The local retention rate $\gamma_t$ is then:
\begin{equation}
    \gamma_t = [\gamma_{\min} + (\gamma_{\max} - \gamma_{\min}) \cdot \hat{H}_t]_{\gamma^{\text{abs}}_{\text{min}}}^{\gamma^{\text{abs}}_{\text{max}}}
    \label{eq:gamma_local_simplified}
\end{equation}
where $\gamma_{\min}, \gamma_{\max}$ are soft bounds, and $[\cdot]_{\text{min}}^{\text{max}}$ denotes clipping.
$\gamma_t$ determines the dynamic GoGI threshold $\tau_t$:
\begin{equation}
    \tau_t = Q_{(1-\gamma_t)\times 100}(\{\mathcal{G}_j | j \in \mathcal{I}_{\text{valid}}\})
    \label{eq:dynamic_threshold}
\end{equation}
where $\mathcal{I}_{\text{valid}}$ excludes non-substantive tokens.

\subsubsection{Adaptive N-Constraint (ANC) for Coherence}
\label{sssec:anc}

To preserve coherence, ANC limits the number of consecutive pruned tokens ($N_t$) based on local complexity, estimated via windowed entropy $\bar{H}_{t, W}$:

\begin{equation}
    \bar{H}_{t, W} = \frac{1}{W} \sum_{j=\max(1, t-\lfloor W/2 \rfloor)}^{\min(m, t+\lfloor W/2 \rfloor)} H_j
    \label{eq:window_entropy}
\end{equation}
\begin{equation}
    \hat{\bar{H}}_{t, W} = \mathcal{M}'(\bar{H}_{t, W}; \theta_{\mathcal{M}'})
\end{equation}

The adaptive constraint $N_t$ is:
\begin{equation}
    \tilde{N}_t = N_{\min} + (N_{\max} - N_{\min}) \cdot (1 - \hat{\bar{H}}_{t, W})
\end{equation}
\begin{equation}
    \label{eq:adaptive_n}
    N_t = \left\lfloor \left[ \tilde{N}_t \right]_{N_{\min}}^{N_{\max}} + 0.5 \right\rfloor
\end{equation}
High local entropy thus yields a smaller $N_t$, enforcing conservative pruning.

\subsection{Adaptive GoGI-Skip Pruning Algorithm}
\label{ssec:pruning_algorithm}

Algorithm~\ref{alg:gogi_skip} integrates GoGI with EDR and ANC into a unified pruning procedure. Tokens are retained if intrinsically important ($\mathcal{G}'_t \ge \tau_t$) or if pruning would violate the ANC constraint ($C_{t-1} + 1 \ge N_t$). This dual-safety mechanism prevents fragmented chains common in pure importance sampling. The resulting compressed sequences $c'_{\text{comp}}$ form the SFT training set $\mathcal{D}_{SFT}$.

\section{Experiments}
\label{sec:experiment}

\begin{table*}[t]
    \centering
    \resizebox{\textwidth}{!}{%
        \begin{tabular}{l|cc|cc|cc|cc}
            \toprule
            \multirow{2}{*}{\raisebox{-0.6ex}{Method Variant}}           & \multicolumn{2}{c|}{AIME 2025}                    & \multicolumn{2}{c|}{AIME 2024} & \multicolumn{2}{c|}{GPQA Diamond}                  & \multicolumn{2}{c}{GSM8K}                                                                                                                                       \\
            \cmidrule(lr){2-3}\cmidrule(lr){4-5}\cmidrule(lr){6-7}\cmidrule(lr){8-9}
                                                      & Acc.                                              & Speedup                        & Acc.                                               & Speedup                   & Acc.                                               & Speedup     & Acc.                                               & Speedup     \\
            \midrule
            Full Method (GoGI+EDR+ANC)                & 5.2                                               & 1.6$\times$                    & 14.1                                               & 1.7$\times$               & 29.3                                               & 1.8$\times$ & 89.9                                               & 1.9$\times$ \\
            \midrule
            w/o ANC (GoGI+EDR only)                   & 4.5$_{\textcolor[HTML]{E53935}{(0.7\downarrow)}}$ & 1.7$\times$                    & 13.6$_{\textcolor[HTML]{E53935}{(0.5\downarrow)}}$ & 1.8$\times$               & 28.5$_{\textcolor[HTML]{E53935}{(0.8\downarrow)}}$ & 1.9$\times$ & 89.0$_{\textcolor[HTML]{E53935}{(0.9\downarrow)}}$ & 2.0$\times$ \\
            w/o EDR (GoGI-Static, $\gamma\approx0.5$) & 4.3$_{\textcolor[HTML]{E53935}{(0.9\downarrow)}}$ & 1.6$\times$                    & 13.5$_{\textcolor[HTML]{E53935}{(0.6\downarrow)}}$ & 1.5$\times$               & 28.6$_{\textcolor[HTML]{E53935}{(0.7\downarrow)}}$ & 1.6$\times$ & 89.5$_{\textcolor[HTML]{E53935}{(0.4\downarrow)}}$ & 1.6$\times$ \\
            w/o ADS (GoGI-Static, $\gamma\approx0.5$) & 4.0$_{\textcolor[HTML]{E53935}{(1.2\downarrow)}}$ & 1.9$\times$                    & 13.0$_{\textcolor[HTML]{E53935}{(1.1\downarrow)}}$ & 1.9$\times$               & 28.0$_{\textcolor[HTML]{E53935}{(1.3\downarrow)}}$ & 1.9$\times$ & 88.5$_{\textcolor[HTML]{E53935}{(1.4\downarrow)}}$ & 1.9$\times$ \\
            w/o GoGI (LLMLingua+ADS)                  & 4.6$_{\textcolor[HTML]{E53935}{(0.6\downarrow)}}$ & 1.6$\times$                    & 13.8$_{\textcolor[HTML]{E53935}{(0.3\downarrow)}}$ & 1.7$\times$               & 28.8$_{\textcolor[HTML]{E53935}{(0.5\downarrow)}}$ & 1.8$\times$ & 89.1$_{\textcolor[HTML]{E53935}{(0.8\downarrow)}}$ & 1.9$\times$ \\
            \bottomrule
        \end{tabular}%
    }
    \caption{\textbf{Component Ablation Analysis on Gemma3-4B.} Identifying the contribution of each module. Removing dynamic regulation (w/o ADS/ANC) causes the maximal performance drop, confirming that static compression cannot accommodate the variable complexity of reasoning chains.}
    \label{tab:ablation_study_final}
\end{table*}

We evaluate Adaptive GoGI-Skip to demonstrate its efficiency and accuracy across benchmarks and model architectures. We further analyze the contribution of core components through ablation studies.

\subsection{Experimental Setup}
\label{ssec:exp_setup}

\paragraph{Models}
Our experiments employ two open-source instruction-tuned model families: Gemma3-Instruct~\cite{team2025gemma} and Qwen2.5-Instruct~\cite{yang2024qwen2}. This selection enables a thorough evaluation of scalability and generalizability.

\paragraph{Training Data}
We sourced samples from the MATH training dataset~\cite{hendrycksmath2021} only, isolating transfer from the effects of broad multi-domain supervision. After standardizing formats and validating parsing, 7,472 valid samples remained. We compress each original CoT trace with Adaptive GoGI-Skip as an SFT set. We then report zero-shot results on AIME, GPQA, and GSM8K across architectures without additional task-specific training.

\paragraph{Baselines}
We compare against CoT compression baselines compatible with the SFT pipeline, including the standard CoT, denoted Original; zero-shot prompting, referred to as Prompting; C3oT, an external model–based compression method~\cite{kang2025c3ot}; Spiritft, which performs perplexity-guided token pruning~\cite{cui2025stepwise}; and TokenSkip, which applies static token skipping at ratios of 0.5 and 0.6~\cite{xia2025tokenskip}.

\paragraph{Implementation Details}
We computed GoGI scores at model-specific optimal layers. ADS hyperparameters were fixed across experiments. Token type-weighting was disabled. All models were fine-tuned using Low-Rank Adaptation (LoRA)~\cite{hu2022lora} on 3 NVIDIA 4090 GPUs. Further details are in the Supplementary Material.

\subsection{Main Results}

We evaluated Adaptive GoGI-Skip's efficiency and accuracy balance. All metrics are averaged over 3 independent fine-tuning runs.

Table~\ref{tab:main_results_gemma4b} shows that Adaptive GoGI-Skip consistently outperforms baselines, achieving $\sim$1.7$\times$ speedups on AIME and 1.9$\times$ on GSM8K. This efficiency stems from dynamic retention: the model retains $\sim$60\% of tokens for complex AIME tasks versus $\sim$45\% for GSM8K. Accuracy remains robust, with negligible drops around 0.4\% on GPQA/AIME'24 and a 0.3\% gain on GSM8K. Even on AIME'25, the 0.9\% drop yields a competitive score. In contrast, static baselines like TokenSkip face a rigid trade-off that high speedup costs significant accuracy, while conservative settings limit speedup. Adaptive GoGI-Skip breaks this compromise, matching original accuracy with aggressive efficiency.

Table~\ref{tab:generalization_models} confirms strong generalization across model scaling and architectures. Efficiency gains scale positively with model size from 1B to 12B. For Gemma3-12B, we achieve a 2.0$\times$ speedup on GSM8K while improving accuracy by 0.3\% relative to Original as the pruned tokens function mainly as reasoning noise. The method works equally well on Qwen2.5, maintaining exact parity on AIME'24 and GPQA while accelerating GSM8K by 1.9$\times$. This consistency indicates that the learned compression policy captures universal features of logical redundancy.

\begin{figure*}[t]
    \centering
    \begin{minipage}{0.52\linewidth}
        \centering
        \includegraphics[width=\linewidth]{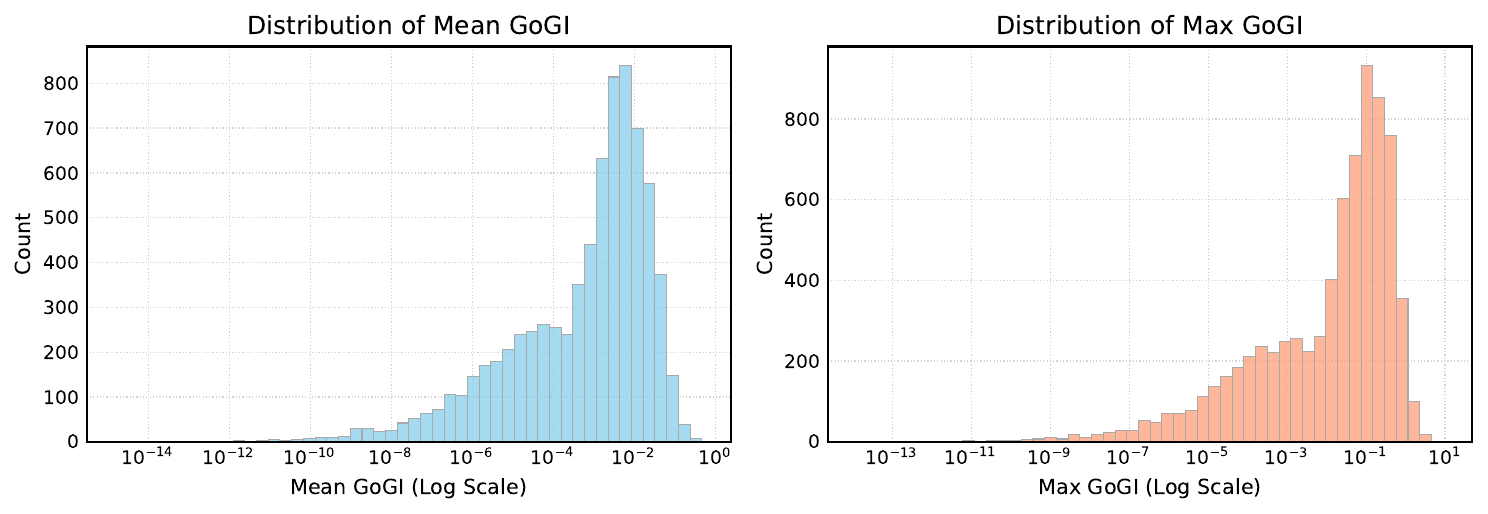}
        \par\vspace{1pt}
        \includegraphics[width=\linewidth]{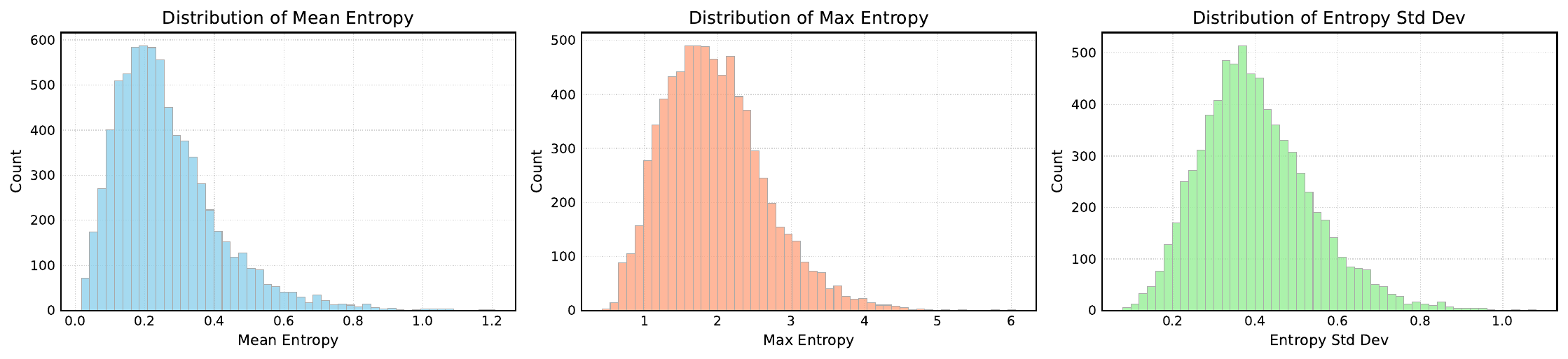}
    \end{minipage}
    \hfill
    \begin{minipage}{0.46\linewidth}
        \centering
        \includegraphics[width=\linewidth]{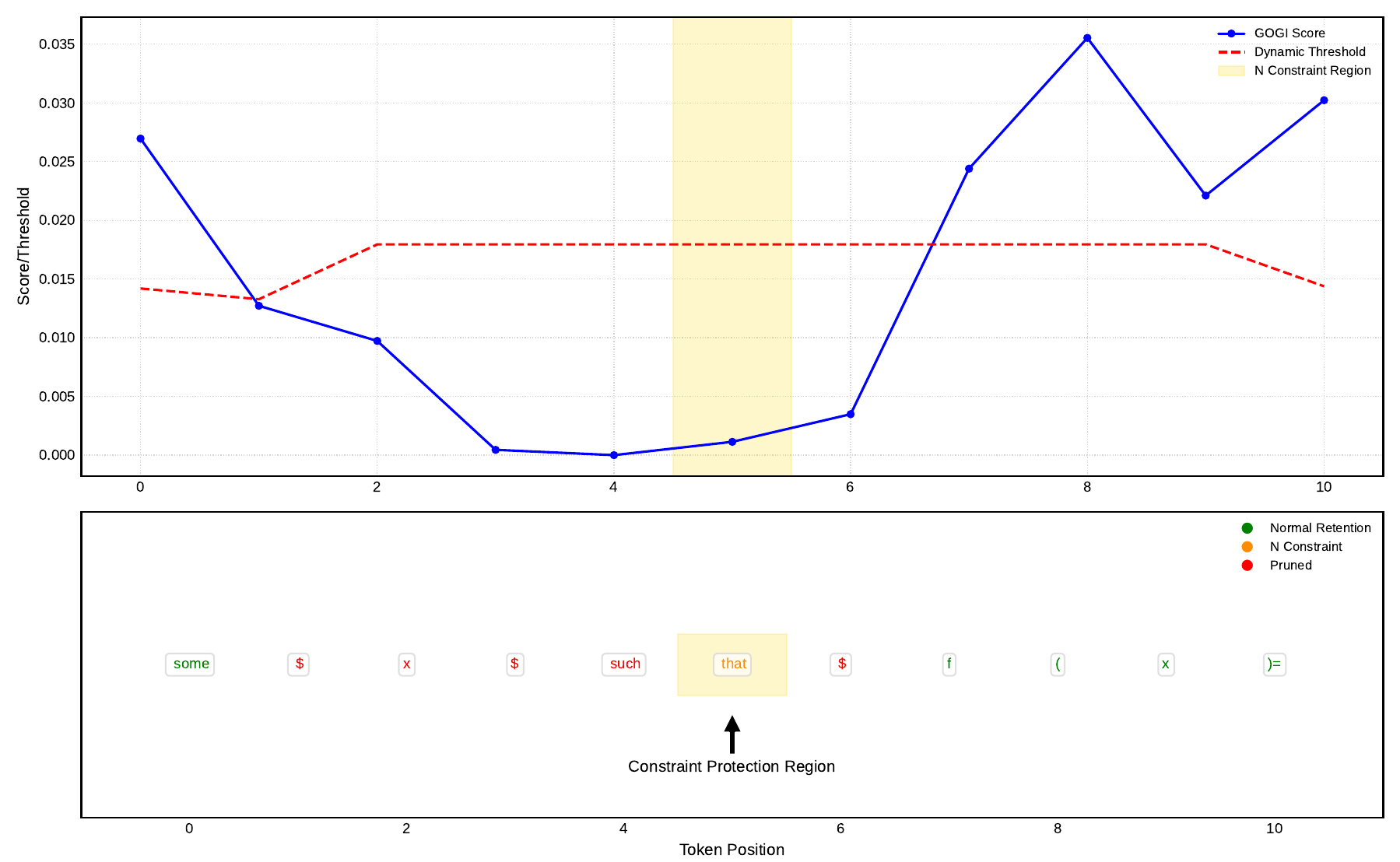}
    \end{minipage}
    \caption{\textbf{Distributional Dynamics of Information.} \textbf{Left}: GoGI scores (top) follow a long-tail distribution, while entropy (bottom) identifies mode switches. \textbf{Right}: ANC dynamically relaxes pruning strictness during entropy spikes, preserving structural coherence at critical junctions.}
    \label{fig:gogi_and_dynamic}
\end{figure*}

Table~\ref{tab:ablation_study_final} quantifies each module's contribution. Removing ADS causes the sharpest accuracy decline despite achieving highest speedup, confirming that static pruning cannot accommodate variable information density. Without ANC, performance drops on harder tasks by 0.7\% on AIME'25, where longer-range dependencies require periodic anchor tokens to maintain logical coherence. The w/o EDR variant underperforms despite matching compression rates since entropy-driven modulation prevents cascading failures at high-entropy branching points. Replacing GoGI with LLMLingua degrades accuracy at comparable speedups, indicating that perplexity-based salience does not reliably recover the steps that drive answer correctness. GoGI selects tokens by their sensitivity to the answer loss, and ADS turns this objective-aligned signal into stable, uncertainty-aware pruning. Together, they yield the best accuracy retention at matched speedups in our ablations.

\subsection{Analysis}
\label{ssec:analysis}
\paragraph{Robustness on Efficiency--Accuracy Frontier.}
Absolute low scores on AIME reflect base model limitations. Even so, Adaptive GoGI-Skip delivers the strongest accuracy retention at comparable speedups across baselines. While baselines such as C3oT and TokenSkip achieve speedups by sacrificing significant accuracy, Adaptive GoGI-Skip preserves performance even at high compression rates. Results on rigorous benchmarks like AIME are particularly telling that they represent robust evidence that our method identifies functionally critical tokens with greater precision than prior SFT approaches. Superior accuracy preservation under these adversarial conditions demonstrates structural alignment with the model's reasoning topology that static heuristic pruning fails to capture.

\paragraph{Effectiveness of Adaptive GoGI-Skip.}
Figure~\ref{fig:gogi_and_dynamic} (left) summarizes the signals that make compression reliable. GoGI scores follow a long-tailed distribution, which indicates that most tokens contribute little to the answer loss. The entropy distribution separates stable segments from uncertain transitions, where pruning errors are more likely to cascade. This separation explains why Adaptive GoGI-Skip prunes aggressively in stable regions while remaining conservative around high-uncertainty steps.

\paragraph{Dynamic Behavior of ADS.}
Figure~\ref{fig:gogi_and_dynamic} (right) visualizes the dynamic adaptability of our strategy. EDR dynamically expands the retention window ($\gamma_t$) during high-entropy segments, allocating compute to ambiguous transitions. Simultaneously, ANC acts as a structural stabilizer. ANC intervenes at critical transitions, such as shifting from calculation to conclusion, preventing the model from skipping logically vital but semantically stable connectors like ``therefore'' and implies. This dual-regulation inversely correlates pruning intensity with local reasoning difficulty, maintaining thinking-optimal density.

\begin{figure}[h]
    \centering
    \includegraphics[width=\linewidth]{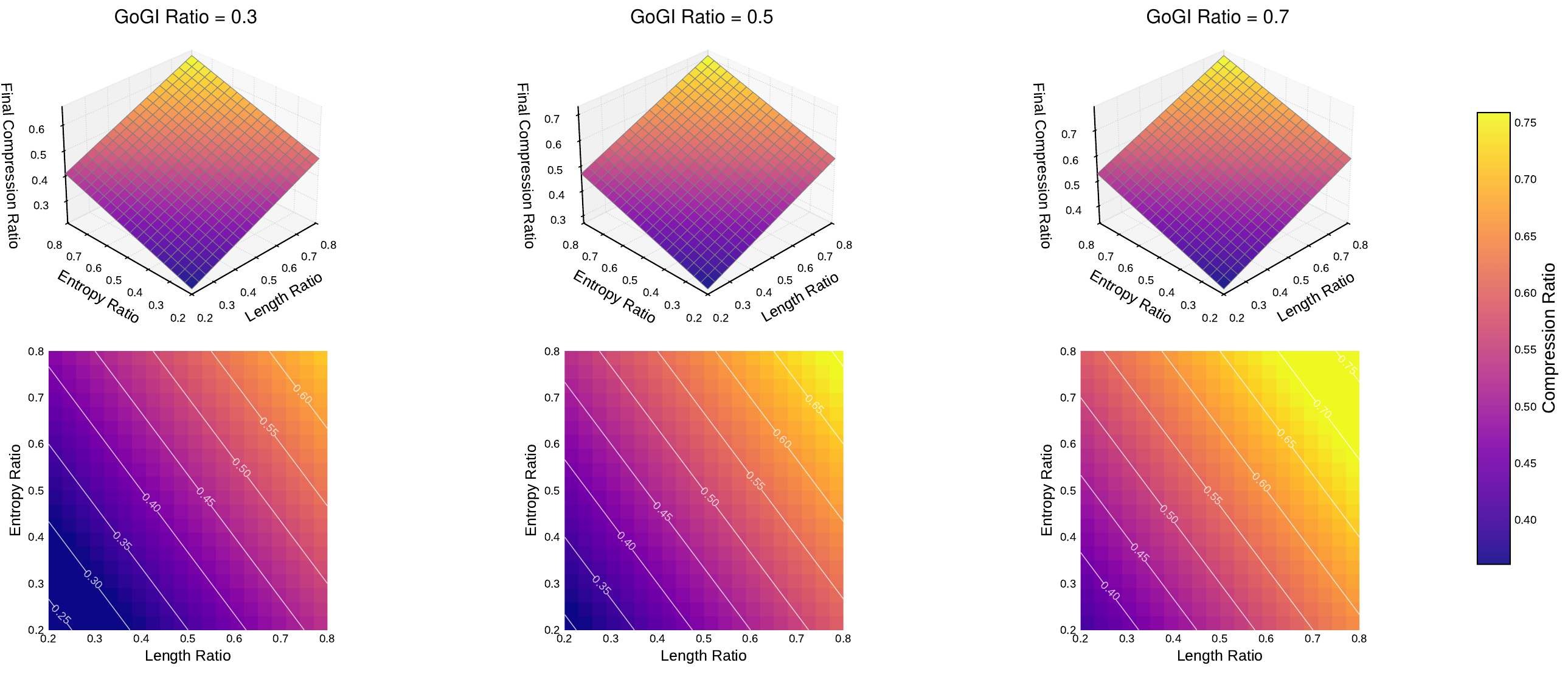}
    \caption{\textbf{Decision Surface of Adaptive Compression.} The joint policy $\gamma(H_t, \mathcal{G}_t)$ illustrates the safety mechanism: high uncertainty forces token retention regardless of GoGI score, while low uncertainty allows aggressive pruning of low-utility tokens.}
    \label{fig:3d_analysis}
\end{figure}

\paragraph{Joint Compression Logic.}
Figure~\ref{fig:3d_analysis} illustrates the 3D visualization which elucidates the joint regulation of compression. We observe a hierarchical interaction that in high-entropy regimes, retention rates remain high regardless of GoGI scores, as EDR enforces safety during uncertainty. Conversely, in low-entropy confident zones, the surface dips significantly, allowing GoGI to aggressively prune tokens with low utility. This pattern confirms that our mechanism effectively prioritizes structural stability over local efficiency when necessary, preventing the blind skipping common in static methods.

\paragraph{Complexity Analysis.}
Adaptive GoGI-Skip reduces the theoretical attention complexity from $O(L^2)$ to $O((1-\bar{\gamma})^2 L^2)$, yielding an approximate $4\times$ reduction in FLOPs for $\bar{\gamma}=0.5$. In practice, since LLM inference is memory-bandwidth bound, the observed $2.0\times$ speedup stems primarily from reduced memory access frequency for weights and KV-caches. Crucially, because all decision parameters are baked into the SFT weights, this acceleration incurs zero runtime overhead: standard decoding simply outputs a shorter, denser sequence.

\begin{figure*}[t]
    \begin{minipage}{0.4\linewidth}
        \centering
        \includegraphics[width=\linewidth]{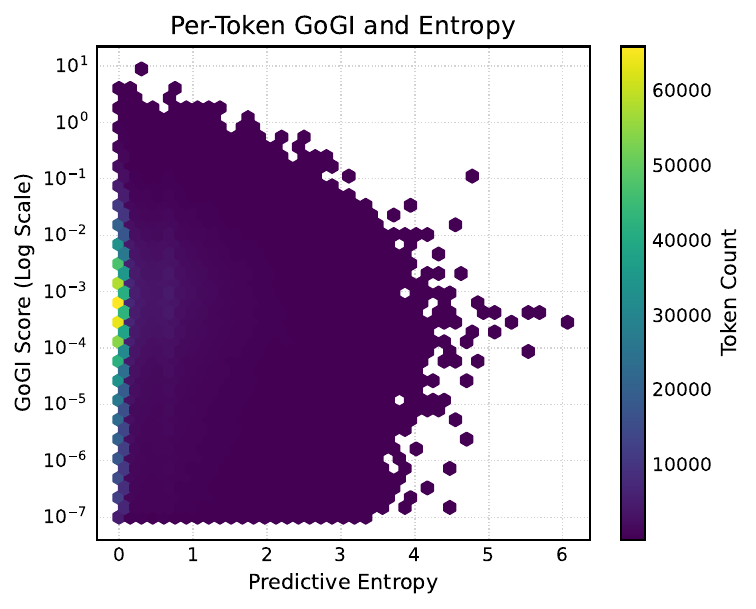}
    \end{minipage}
    \begin{minipage}{0.58\linewidth}
        \centering
        \includegraphics[width=\linewidth]{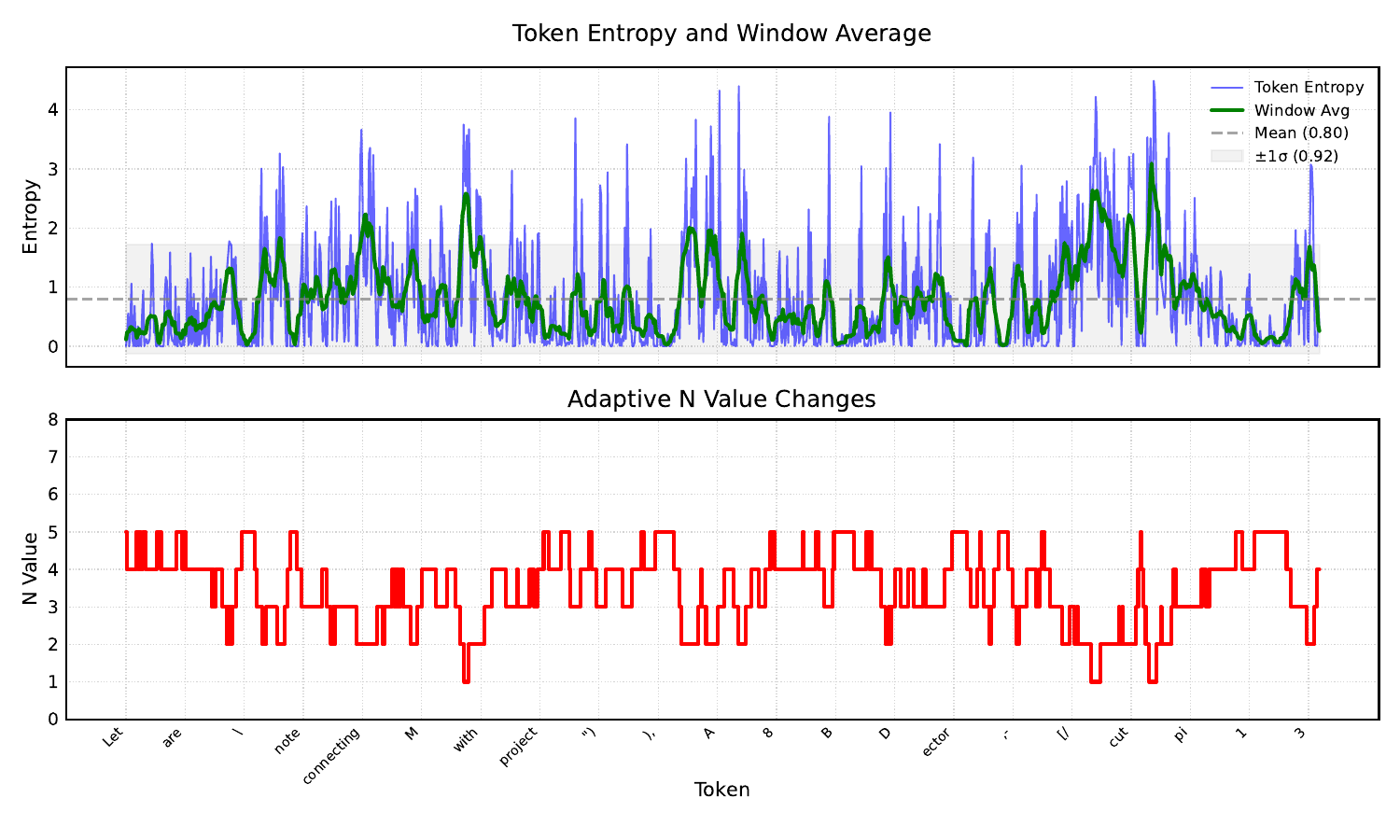}
    \end{minipage}
    \caption{\textbf{Orthogonality and Regulation.} \textbf{Left:} The lack of correlation between GoGI and Entropy implies they capture distinct feature dimensions (Teleological vs. Epistemic). \textbf{Right:} The ANC windowing acts as a low-pass filter on skipping decisions, smoothing the compression rate to prevent fragmented chains.}
    \label{fig:discussion}
\end{figure*}
\section{Discussion}
\label{sec:discussion}

Our experiments validate that Adaptive GoGI-Skip addresses the efficiency--accuracy trade-off in CoT reasoning. Here we synthesize the implications.

\subsection{Implications and Insights}

\paragraph{Beyond uncertainty-based proxies.}
\label{ssec:discuss_gogi}
The near-zero correlation between GoGI scores and predictive entropy (Figure~\ref{fig:discussion}, left) challenges reliance on uncertainty as an importance proxy. Statistical surprise ($H_t$) signals branching points or ambiguity but cannot distinguish necessary functional steps from linguistic idiosyncrasies. As Jain and Wallace~\cite{jain2019attention} note, surface-level attention patterns often diverge from true model reasoning; uncertainty similarly fails to predict valid pruning candidates. By anchoring importance to the answer-loss gradient, GoGI enables aggressive pruning of semantically distinct but functionally redundant tokens, a capability missed by perplexity-based methods that conflate surprising with important.

\paragraph{The case for adaptivity.}
\label{ssec:discuss_ads}
Rigid trade-offs in static baselines underscore the profound heterogeneity of reasoning steps: some require deep computational verification; others are routine syntactic bridges. ADS decouples token count from task difficulty, echoing Adaptive Computation Time principles~\cite{graves2016adaptive}. As Figure~\ref{fig:discussion} (right) shows, ANC preserves local structural integrity in high-entropy regions even when GoGI signals are low. This synergy enables heavy compression of routine logic while retaining full fidelity for crux moments, approximating thinking-optimal density essential for sustainable Green AI~\cite{schwartz2020green}.

\paragraph{Compatibility with RL-based reasoners.}
\label{ssec:discuss_sft}
RL-based reasoning models such as DeepSeek-R1~\cite{guo2025deepseek} achieve strong performance but produce increasingly verbose traces~\cite{yeo2025demystifying}. Adaptive GoGI-Skip offers a natural post-processing choice: it compresses RL-generated chains without modifying the base model or reward function. This decoupled design preserves RL's performance benefits while recovering inference efficiency, making our method applicable to any verbose reasoner regardless of its training paradigm.

\subsection{Broader Impact}
\label{ssec:impact}
Our work advances practical AI deployment along two axes. Detailed CoT generation multiplies the energy cost per query, and reducing token count by $\sim$45\% lowers kilowatt-hour consumption proportionally, supporting Green AI initiatives~\cite{luccioni2025hungry,patterson2025energy}. Speedups reduce end-to-end latency, increase system throughput, and improve resource utilization. This supports higher concurrency and makes it easier to deploy more capable models under fixed infrastructure constraints.

\subsection{Limitations}
\label{ssec:limitations}
Since GoGI computation requires direct access to model parameters and gradients during data preparation, the method applies to all open models with full parameter access, but it does not extend to models that are only available through closed-source APIs. While inference accelerates, the offline GoGI score calculation incurs a one-time preprocessing cost that scales with the training corpus size, yet this cost amortizes over repeated deployments of the accelerated model.
\section{Conclusion}
\label{sec:conclusion}
Adaptive GoGI-Skip preserves CoT reasoning while enabling efficient deployment. By unifying goal-gradient importance with dynamic uncertainty regulation, it achieves a principled compression that minimizes redundancy without sacrificing accuracy. Trained on only 7,472 MATH traces, the policy transfers zero-shot across benchmarks and architectures, delivering $>$45\% token reduction and up to 2.0$\times$ speedup with negligible accuracy loss. These results show that structural redundancy in reasoning is a learnable, task-agnostic property distinct from semantics. As RL-driven scaling yields ever-longer traces, GoGI-Skip isolates core deductive signals from superfluous content and embeds compact reasoning into model weights via standard SFT, setting a new baseline for resource-efficient inference.

\section*{Acknowledgments}
I am grateful to Professor Ben Wang and Professor Shuifa Sun from Hangzhou Normal University for their guidance, encouragement, and continued support throughout the course of this work.

\bibliographystyle{named}
\bibliography{ijcai26}

\newpage
\appendix
\setcounter{table}{0}
\renewcommand{\thetable}{\arabic{table}}
\setcounter{figure}{0}
\renewcommand{\thefigure}{\arabic{figure}}
\setcounter{equation}{0}
\renewcommand{\theequation}{\arabic{equation}}
\part*{Appendices}
\section{Detailed Experimental Setup}
\label{app:setup_details}

This appendix provides comprehensive details regarding the experimental setup, supplementing \S \ref{ssec:exp_setup}.

\subsection{Training Data Source and Pre-processing}
\label{app:data_preprocessing}

\paragraph{Data Source.}
We sourced training data from the MATH dataset's canonical subset (7,500 pairs). Each pair comprises a problem statement, a reference Chain-of-Thought (CoT) solution, and the ground truth answer.

\paragraph{CoT and Answer Extraction.}
We extracted CoT reasoning from the solution field using regex pattern matching delimited by the standard $\boxed{answer}$ format, thereby isolating the reasoning path from the final solution.

\paragraph{Validation Protocol.}
We verified mathematical equivalence between extracted answers and ground truth values via a symbolic mathematics library, retaining only samples with strict symbolic equivalence. We tokenized all CoTs using model-specific tokenizers. We excluded samples exceeding 1,500 tokens or those causing computational infeasibility during GoGI calculation. This filtering yielded 7,472 valid samples for training.

\subsection{GoGI Target Layer Selection Justification}
\label{app:layer_contribution}

\paragraph{Analysis Methodology and Rationale.}
The target layer $l^*$ for GoGI calculation governs importance signal quality. Because computing GoGI scores across all layers is intractable, we maximized efficiency by analyzing layer contribution to identify a representative target layer $l^*$. This metric quantifies, for each layer $l$, the average L1 norm of gradients propagated from the top-1 prediction logit (at each valid CoT position $t$) back to the input activation $\mathbf{x}_t^l$ of that layer's MLP down-projection module (Algorithm \ref{alg:layer_contribution}). This reflects the layer's local processing intensity. We hypothesized that layers exhibiting high contribution are crucial for information integration and contain intermediate representations $\mathbf{h}_t^l$ sensitive to the final answer loss. This proxy significantly reduces computational cost while identifying promising candidates for $l^*$.

\begin{algorithm}[h!]
    \caption{Layer Contribution Calculation (Single Sample)}
    \label{alg:layer_contribution}
    \begin{algorithmic}[1]
        \renewcommand{\algorithmicrequire}{\textbf{Input:}}
        \renewcommand{\algorithmicensure}{\textbf{Output:}}
        \REQUIRE Model $M_\theta$, Tokenizer $T$, Sample $s$, CoT IDs $c_{ids}$
        \ENSURE Per-layer contributions $Contrib_s = \{l: \bar{g}_l\}$
        \STATE $L \gets$ number of layers in $M_\theta$
        \STATE Register hook on MLP down\_proj for all layers
        \STATE $ids \gets T(\text{Format}(s))$
        \STATE $\mathcal{R} \gets \text{LocateCoT}(c_{ids}, ids)$ \COMMENT{CoT range}
        \IF{$\mathcal{R} = \emptyset$}
        \RETURN $\emptyset$
        \ENDIF
        \STATE $\mathbf{Z} \gets M_\theta(ids)$ \COMMENT{Logits}
        \FOR{$t \in \mathcal{R}$}
        \IF{$t < |\mathbf{Z}| - 1$}
        \STATE $M_\theta.\text{zero\_grad}()$
        \STATE $v^* \gets \arg\max \mathbf{Z}[t]$
        \STATE Backprop from $\mathbf{Z}[t, v^*]$ with retain\_graph
        \FOR{$l = 0$ \TO $L-1$}
        \STATE $G_l[t] \gets \|\nabla_l\|_1$ \COMMENT{L1 norm of hooked grad}
        \ENDFOR
        \ENDIF
        \ENDFOR
        \FOR{$l = 0$ \TO $L-1$}
        \STATE $Contrib_s[l] \gets \text{mean}(G_l)$
        \ENDFOR
        \RETURN $Contrib_s$
    \end{algorithmic}
\end{algorithm}

\begin{figure*}[t]
    \centering
    \includegraphics[width=\linewidth]{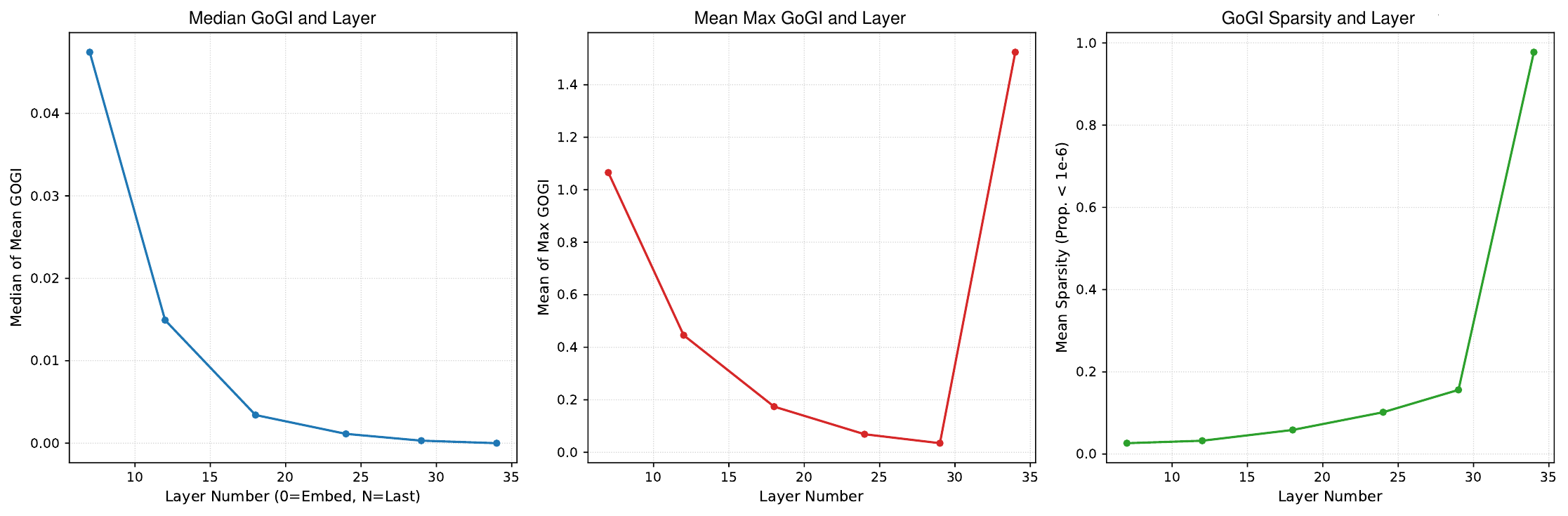}
    \caption{\textbf{GoGI target layer selection analysis (Gemma3-4B-Instruct).} \textbf{Left:} Layer-wise gradient contribution to logits. \textbf{Right:} GoGI statistics across layers. L23 was selected for its strong overall contribution.}
    \label{fig:layer_contribution}
\end{figure*}

\begin{figure}
    \centering
    \includegraphics[width=\linewidth]{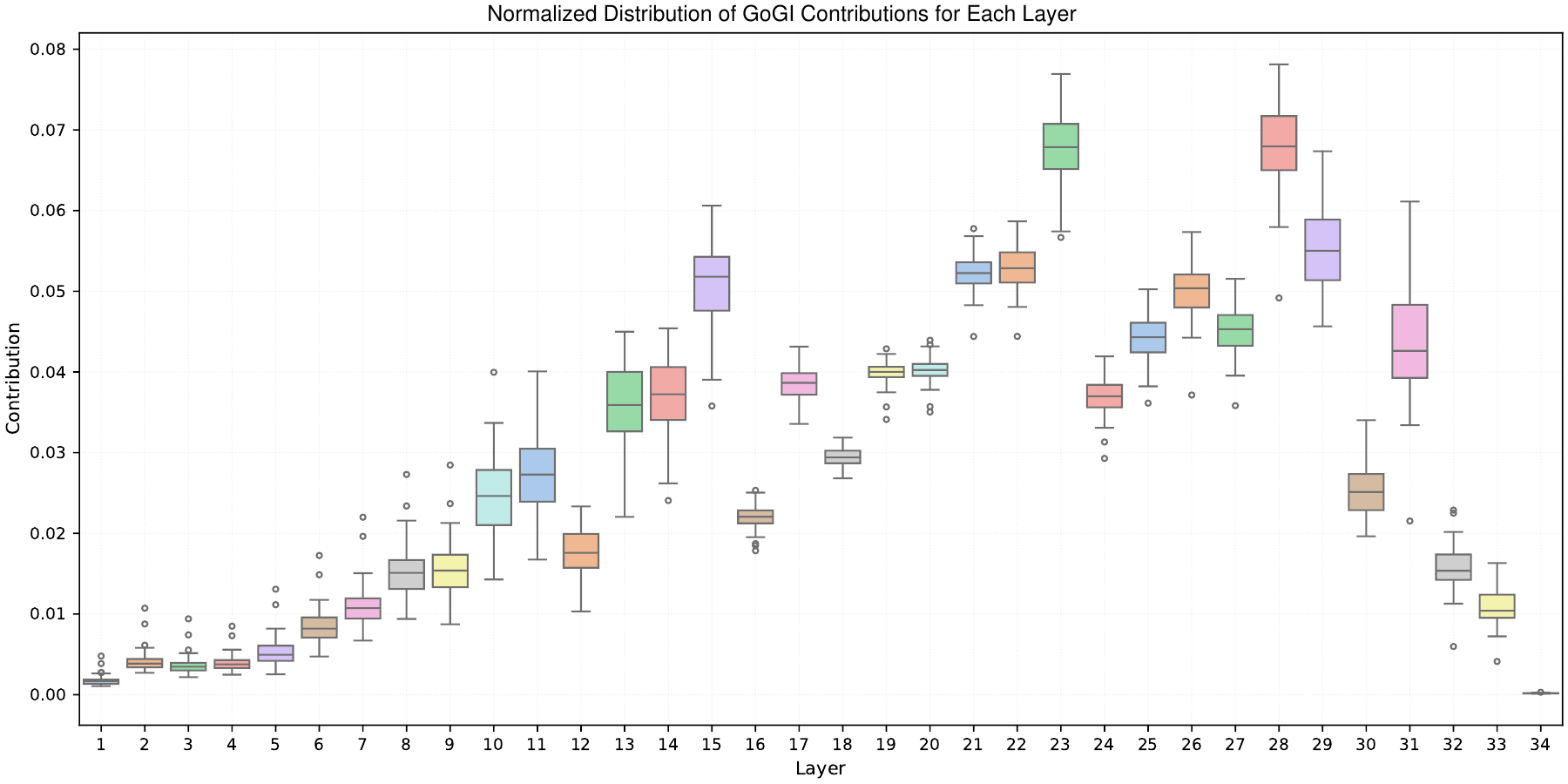}
    \caption{\textbf{Layer contribution variability analysis (Gemma3-4B-Instruct).}}
    \label{fig:layer_contribution_variability_appendix}
\end{figure}

\begin{figure*}[t]
    \centering
    \begin{minipage}{0.49\linewidth}
        \centering
        \includegraphics[width=\linewidth]{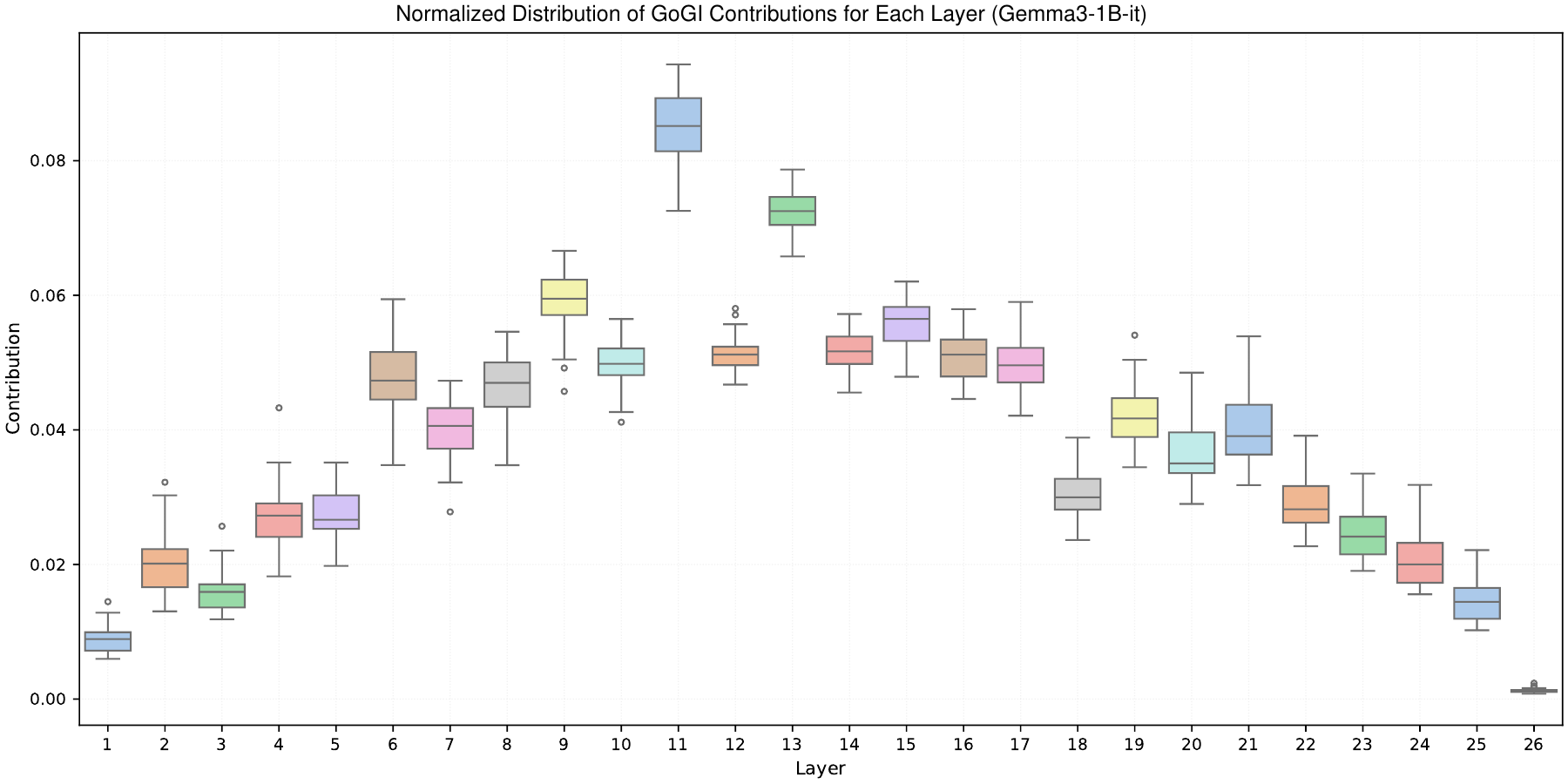}
        \includegraphics[width=\linewidth]{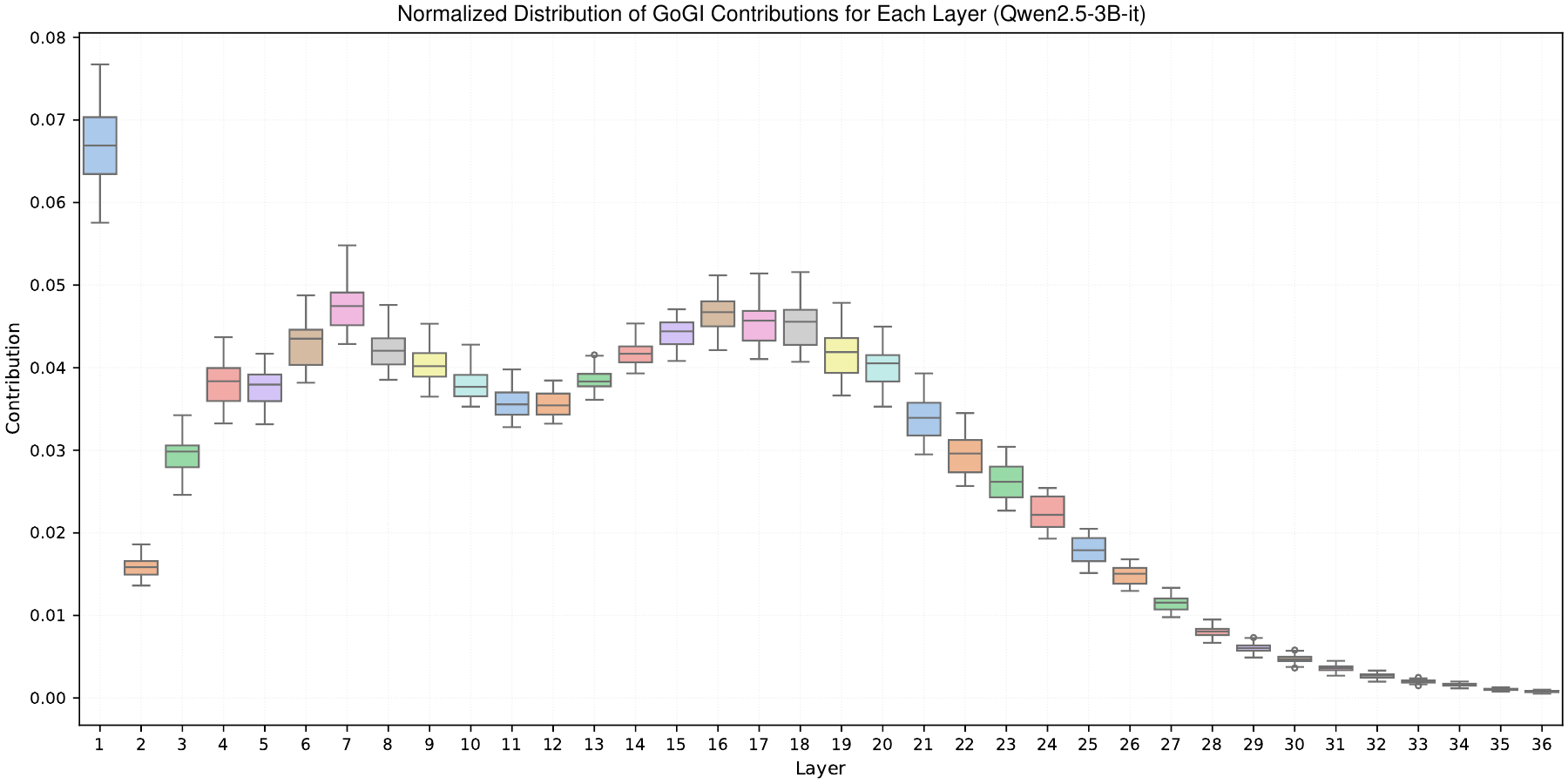}
    \end{minipage}
    \begin{minipage}{0.49\linewidth}
        \centering
        \includegraphics[width=\linewidth]{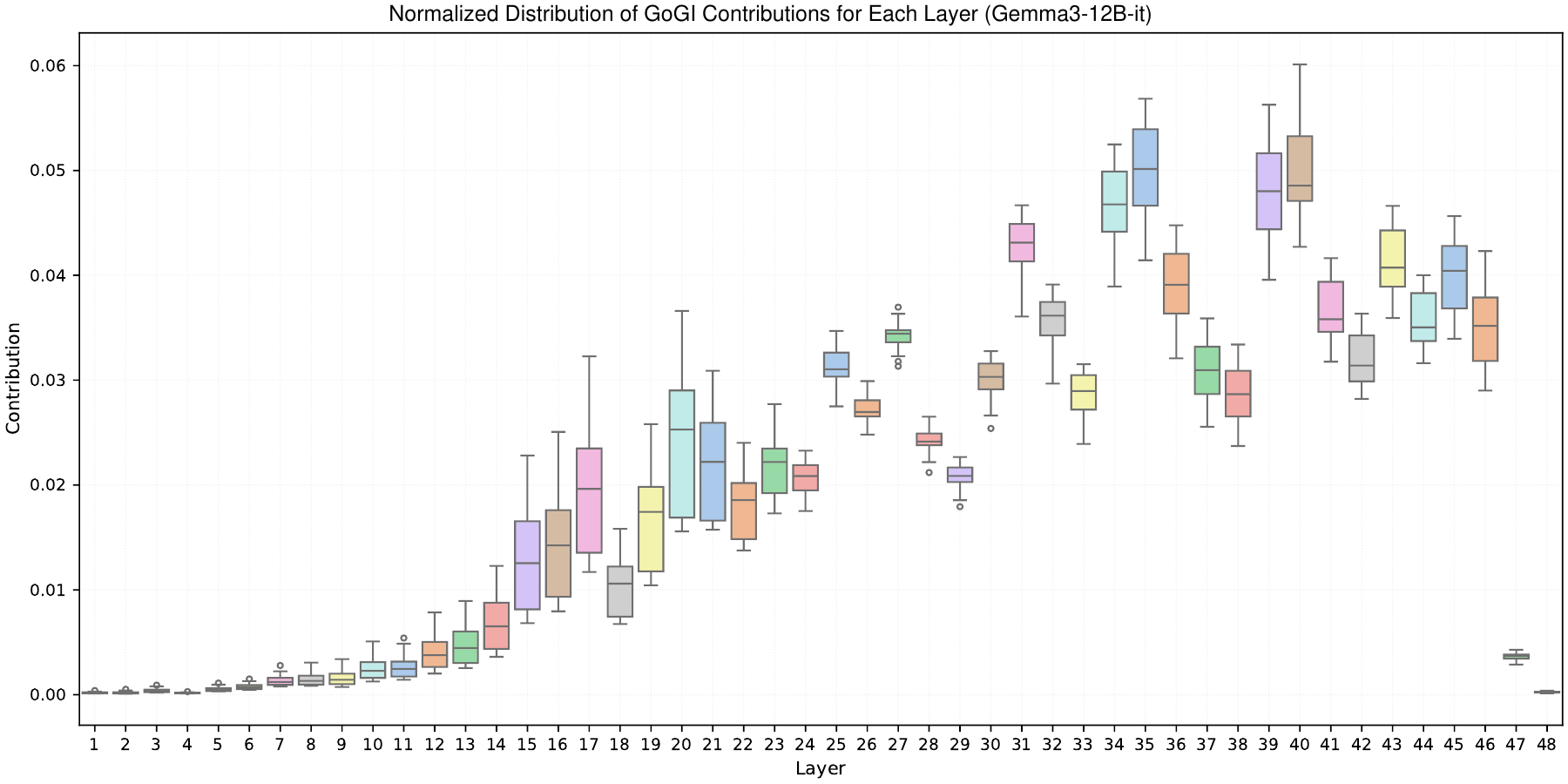}
        \includegraphics[width=\linewidth]{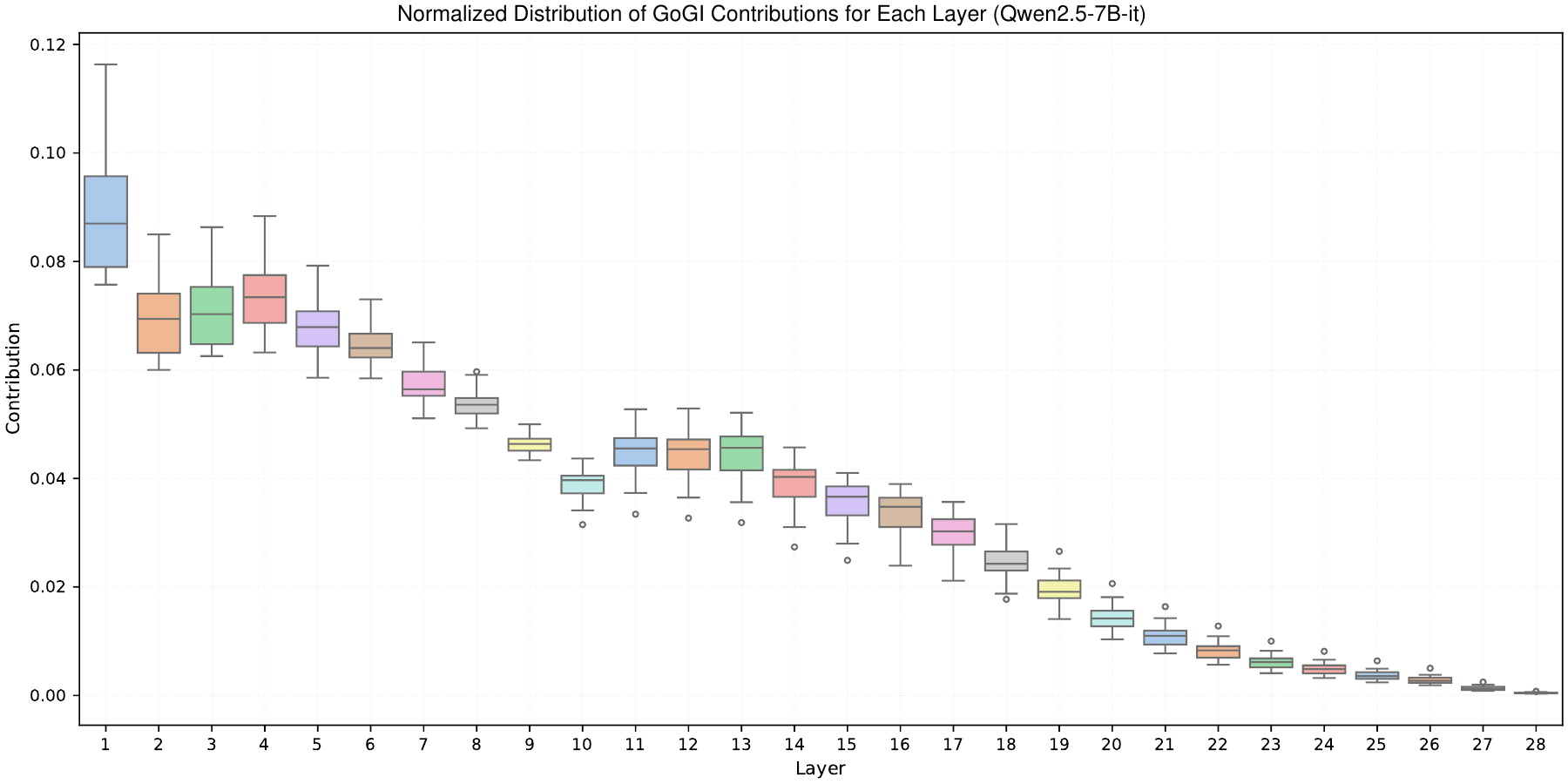}
    \end{minipage}
    \caption{\textbf{Normalized gradient contribution distributions.} \textbf{Top:} Gemma3-1B and 12B. \textbf{Bottom:} Qwen2.5-3B and 7B.}
    \label{fig:appendix_layer_contribution_all}
\end{figure*}

\paragraph{Cross-Model Observations.}
Figure \ref{fig:appendix_layer_contribution_all} illustrates layer-wise contribution distributions, exhibiting significant architectural diversity. Gemma3 models (1B, 4B, 12B) show increasing contributions peaking in middle-to-late stages. In contrast, Qwen2.5-7B-Instruct shows an early peak followed by steep decline, while Qwen2.5-3B-Instruct displays a broader peak around L17-18.
This heterogeneity implies that no universal optimum relative depth for $l^*$ exists; models likely employ different layer specialization strategies. However, preliminary experiments indicated that Adaptive GoGI-Skip performance is robust to $l^*$ selection, provided the layer has high contribution. Selecting near-zero contribution layers degraded performance, but variants among high-contribution layers yielded comparable results. This robustness stems from: (1) Preserved relative token importance rankings across mid-to-late layers; (2) Compensation by ADS mechanisms (EDR and ANC); and (3) Adaptation during SFT.

\paragraph{Selection Principle.}
Given this robustness and variance, we adopted a heuristic selection principle: identify layer(s) exhibiting the latest visually significant peak or plateau in gradient contribution (Figure \ref{fig:appendix_layer_contribution_all}). We prioritized later layers within this set to capture integrated semantic understanding before output saturation. Table \ref{tab:layer_selection} details the selected target layers $l^*$.

\begin{table}[h]
    \centering
    \caption{\textbf{Target Layer ($l^*$) Selection.} Selections differ by model architecture based on gradient contribution peaks.}
    \label{tab:layer_selection}
    \begin{tabular}{lcc}
        \toprule
        Model Name          & Layer Size & Selected Layer ($l^*$) \\
        \midrule
        Gemma3-1B-Instruct  & 26         & 18                     \\
        Gemma3-4B-Instruct  & 34         & 23                     \\
        Gemma3-12B-Instruct & 48         & 35                     \\
        Qwen2.5-3B-Instruct & 36         & 18                     \\
        Qwen2.5-7B-Instruct & 40         & 27                     \\
        \bottomrule
    \end{tabular}
\end{table}

\subsection{Adaptive GoGI-Skip Hyperparameters}
\label{app:ads_hyperparams}

To enhance robustness across datasets, we developed an optional Adaptive Parameter Tuner utilized during initialization.

\paragraph{Motivation}
Optimal ADS hyperparameters (e.g., base retention $\gamma_{\text{base}}$, entropy scales $s_\gamma, s_N$) depend on dataset statistics. The Tuner adjusts these automatically.

The Tuner operates in three steps:
\begin{enumerate}
    \item \textbf{Feature Extraction:} Computes statistical features (e.g., average CoT length, entropy moments) from a data subset.
    \item \textbf{Target Ratio Estimation:} Estimates a global compression ratio $\gamma_{target}$ via a heuristic function based on entropy distribution (higher variance suggests conservative compression).
    \item \textbf{Parameter Update:} Adjusts hyperparameters based on $\gamma_{target}$.
\end{enumerate}

Table \ref{tab:ads_hyperparams_final} lists default ADS hyperparameters.

\begin{table}[h]
    \centering
    \caption{\textbf{Adaptive GoGI-Skip Hyperparameters.} Defaults used unless Tuner is activated for dataset-specific adjustment.}
    \label{tab:ads_hyperparams_final}
    \begin{tabular}{lcc}
        \toprule
        Parameter             & Symbol                        & Default    \\
        \midrule
        \multicolumn{3}{l}{\textbf{EDR Regulation}}                        \\
        Min Retention Rate    & $\gamma_{\min}$               & 0.2        \\
        Max Retention Rate    & $\gamma_{\max}$               & 0.8        \\
        Base Gamma            & $\gamma_{\text{base}}$        & 0.6        \\
        Entropy Delta         & $\delta_e$                    & 0.3        \\
        Entropy Scale Factor  & $s_{\gamma}$                  & 1.8        \\
        Entropy Mapping Mode  & $mode$                        & auto       \\
        Entropy Median        & $H_{\text{median}}$           & Estimated  \\
        Entropy Std Dev       & $H_{\text{std}}$              & Estimated  \\
        \midrule
        \multicolumn{3}{l}{\textbf{ANC Coherence}}                         \\
        Min Consecutive Skips & $N_{\min}$                    & 1          \\
        Max Consecutive Skips & $N_{\max}$                    & 9          \\
        Entropy Window Size   & $W$                           & 9          \\
        Entropy Scale Factor  & $s_{N}$                       & 1.8        \\
        \midrule
        \multicolumn{3}{l}{\textbf{Token Filtering}}                       \\
        Ignored Token IDs     & $\mathcal{I}_{\text{ignore}}$ & {Space ID} \\
        \bottomrule
    \end{tabular}
\end{table}

\paragraph{Token Filtering ($\mathcal{I}_{\text{valid}}$).}
Dynamic GoGI threshold $\tau_t$ (Eq. \ref{eq:dynamic_threshold}) calculation excludes space tokens by default to prevent skewing percentile statistics.

\paragraph{Entropy Mapping ($\mathcal{M}, \mathcal{M}'$).}
Mapping functions normalize entropy values using global statistics ($H_{\text{median}}, H_{\text{std}}$) and apply a transformation (auto, sigmoid, tanh, gaussian, piecewise). 'Auto' mode selects transformations based on entropy deviation $H_{\text{std}}$ for robustness.

\subsection{Hardware and Software}
\label{app:hardware_software}

\paragraph{Computational Infrastructure.}
We executed experiments on a server with 3$\times$ NVIDIA GeForce RTX 4090 GPUs (24GB VRAM each).

\paragraph{Software Environment.}
The stack included Python 3.12.9, PyTorch 2.6, Transformers 4.51.3, and CUDA 12.4.

\paragraph{GoGI Computation Cost.}
Because GoGI requiring distinct sample-wise backpropagation, parallelization was inefficient on our hardware. We computed GoGI scores on a single RTX 4090 (unquantized). Offline calculation for 7,472 samples (Gemma3-4B) required 12 GPU hours. Dynamic compression (entropy + pruning) took 16 GPU hours. Storing pre-estimated entropy statistics ($H_{\text{median}}, H_{\text{std}}$) avoided re-computation. This one-time cost is comparable to standard RLHF optimization budgets.

\subsection{SFT Hyperparameters}
\label{app:sft_hyperparams}

We fine-tuned all methods using LoRA (Table \ref{tab:sft_hyperparams}).

\begin{table}[t]
    \centering
    \caption{\textbf{LoRA SFT Hyperparameters.} We utilized PEFT with LoRA for all models. Hyperparameters were consistent across methods, with rank and alpha adjusted based on model size.}
    \label{tab:sft_hyperparams}
    \begin{tabular}{ll}
        \toprule
        Parameter             & Value(s)                                       \\
        \midrule
        LoRA Rank ($r$)       & 16 (1B/3B), 32 (4B/7B/12B)                     \\
        LoRA Alpha ($\alpha$) & 32 (1B/3B), 64 (4B/7B/12B)                     \\
        LoRA Dropout          & 0.05                                           \\
        Target Modules        & $q_{proj}$, $k_{proj}$, $v_{proj}$, $o_{proj}$ \\
        Learning Rate         & 2e-5                                           \\
        Batch Size            & ~64-128 (effective)                            \\
        Epochs                & 1-2                                            \\
        LR Scheduler          & cosine (warmup 0.03)                           \\
        Optimizer             & AdamW                                          \\
        Precision             & BF16                                           \\
        \bottomrule
    \end{tabular}
\end{table}

\subsection{Latency Measurement Protocol}

\paragraph{Configuration.}
We measured latency on a single NVIDIA GeForce RTX 4090 (24GB) to ensure hardware consistency.

\paragraph{Context.}
We used a batch size of 1 to reflect interactive reasoning and employed greedy decoding (temp=0) for deterministic output.

\paragraph{Procedure.}
We performed 100 warm-up runs before measuring. We calculated average per-sample latency (end-to-end inference time) over the test set, averaged across 2-3 passes.

\paragraph{Speedup.}
Speedup is defined as:
\begin{equation}
    \label{eq:speedup}
    \text{Speedup}_{\text{Method}} = \frac{\text{AvgLatency}_{\text{Original}}}{\text{AvgLatency}_{\text{Method}}}
\end{equation}
\section{Baseline Implementation Details}
\label{app:baseline_details}

This appendix details implementation specifications for the baseline methods used in Section \ref{sec:experiment}. To ensure rigor, we trained all SFT-based baselines (C3oT, Spiritft, TokenSkip) on compressed CoT data generated using their respective methodologies. We derived this data from the same 7,472 validated MATH training samples used for Adaptive GoGI-Skip, facilitating direct strategy comparison.

For C3oT, we generated compressed CoTs via greedy decoding with the compressor model (Gemma3-4B-Instruct). For Spiritft and TokenSkip, we applied their respective offline pruning algorithms to original CoTs. We then fine-tuned all target models (Gemma3 1B/4B/12B, Qwen2.5 3B/7B) on these datasets (comprising $\langle$problem, compressed CoT, answer$\rangle$ triplets) via LoRA SFT, following the identical procedure (Appendix \ref{app:sft_hyperparams}) used for Adaptive GoGI-Skip.

\subsection{C3oT}
\label{app:c3ot_details}

\begin{tcolorbox}[
        title=C3oT Compression Prompt Template,
        colbacktitle=mytitleblue,
        coltitle=white,
        colframe=mytitleblue,
        colback=white,
        arc=1.5mm,
        boxrule=0.7pt,
        left=2mm,
        right=2mm,
        top=1mm,
        bottom=1mm,
        fonttitle=\bfseries\small
    ]
    \begin{lstlisting}
You will be given a detailed step-by-step reasoning process used to solve a math problem. Your task is to significantly compress this reasoning process while preserving all critical logical steps, and the final conclusion necessary to arrive at the correct answer. Eliminate redundant explanations, repetitive phrasing, trivial arithmetic steps, and verbose language. The compressed reasoning should be clear, logically sound, and contain enough information for someone knowledgeable to verify the solution path. Output ONLY the compressed reasoning process.

## Original Reasoning:
{original cot}

## Compressed Reasoning:
\end{lstlisting}
\end{tcolorbox}

\paragraph{C3oT.} To benchmark against language model-guided compression without proprietary APIs, we implemented a self-compression strategy inspired by C3oT. We used our primary analysis model (\texttt{google/gemma-3-4b-it}) as the compression agent. For each of the 7,472 validated samples $(problem, original\_cot, answer)$, we instructed the model to generate a condensed version of the original CoT using the following prompt:

\subsection{Prompting}
\label{app:prompting_details}

\begin{tcolorbox}[
        title=Prompting Compression Prompt Template,
        colbacktitle=mytitleblue,
        coltitle=white,
        colframe=mytitleblue,
        colback=white,
        arc=1.5mm,
        boxrule=0.7pt,
        left=2mm,
        right=2mm,
        top=1mm,
        bottom=1mm,
        fonttitle=\bfseries\small
    ]
    \begin{lstlisting}
Please provide a concise step-by-step solution to the following problem. Focus only on the essential calculations and logic needed to reach the answer. Avoid verbose explanations or stating obvious facts. Put your final answer within \boxed{answer}.

## Problem:
{problem}

## Solution:
\end{lstlisting}
\end{tcolorbox}

\paragraph{Prompting.} This baseline uses a zero-shot approach without fine-tuning. We directly instructed base models during inference to generate concise reasoning traces. We experimented with prompts targeting conciseness; the optimal template appears below. We prepended this prompt to the standard problem input.

While this zero-shot approach occasionally reduced verbosity, it frequently caused incomplete reasoning or accuracy degradation, highlighting the limitations of relying solely on instructions for complex CoT modification.

\subsection{Spiritft}
\label{app:spiritft_details}

\paragraph{Spiritft.} To evaluate predictability-based compression against goal-oriented importance, we implemented a static pruning baseline inspired by Spiritft, adapted for offline SFT data generation. Spiritft identifies expendable steps by analyzing perplexity changes upon removal. We implemented the following stages:

\begin{enumerate}
    \item \textbf{Reasoning Segmentation:} We decomposed each source CoT into discrete steps ($s_1, s_2, \dots, s_N$), primarily using sentence boundaries and heuristics for mathematical notation.

    \item \textbf{Step Importance Quantification:} We quantified step importance by measuring the impact of omission on model processing fluency. For each step $s_j$, we constructed a pruned CoT $c'_j = s_1\dots s_{j-1}s_{j+1}\dots s_N$. We calculated the increase in average negative log-likelihood ($\Delta \text{NLL}(s_j)$) when the base model processed $c'_j$ versus the original $c$. Lower $\Delta \text{NLL}$ signifies lower importance.

    \item \textbf{Retention-Based Pruning:} We retained the $\lceil \gamma \cdot N \rceil$ steps exhibiting the highest $\Delta \text{NLL}$. We calibrated $\gamma$ to match average compression ratios of TokenSkip variants (0.5 and 0.6) for equitable comparison.

    \item \textbf{Sequence Reconstruction:} We concatenated retained steps in original order. We omitted semantic merging to isolate the effectiveness of perplexity-based importance assessment.
\end{enumerate}

We trained target models on the resulting dataset (comprising $\langle$problem, perplexity-pruned CoT, answer$\rangle$ triplets) via LoRA SFT. This enabled valid comparison between GoGI-based importance and established predictability metrics.

\subsection{TokenSkip}
\label{app:tokenskip_details}
The TokenSkip SFT baseline required offline generation of training data compressed to specific retention ratios ($\gamma=0.5$ and $\gamma=0.6$). We generated datasets via the LLMLingua library:
\begin{enumerate}
    \item We initialized \texttt{llmlingua.PromptCompressor} with \texttt{llmlingua-2-xlm-roberta-large}.
    \item For each original CoT, we used \texttt{compress\_prompt} to generate compressed text.
    \item We set the target \texttt{rate} to $\gamma$, using derived parameters to preserve essential operators.
    \item The \texttt{compressed\_prompt} output served as the compressed CoT for SFT.
\end{enumerate}
This procedure employed LLMLingua to create statics compressed datasets based on internal importance metrics.
\section{Detailed Token Retention Analysis by Functional Type}
\label{app:token_retention_analysis}

\begin{table*}[t]
    \centering
    \caption{Average Token Retention Rate (\%) by Functional Type on the MATH Test dataset (Model: Gemma3-4B-Instruct).}
    \label{tab:appendix_retention_detail}
    \resizebox{\textwidth}{!}{%
        \begin{tabular}{lcccccc|c}
            \toprule
            Method                      & Numerals & Operators & Symbols & Formatting & Connectives & General & Overall \\
            \midrule
            Prompting                   & 95.5     & 94.2      & 93.7    & 88.3       & 85.9        & 82.5    & 0.91    \\
            C3ot                        & 84.1     & 82.8      & 79.5    & 63.4       & 58.2        & 48.6    & 0.67    \\
            Spiritft                    & 80.3     & 78.5      & 75.8    & 56.2       & 51.6        & 40.8    & 0.65    \\
            TokenSkip ($\gamma=0.6$)    & 81.3     & 79.2      & 76.5    & 58.0       & 53.1        & 38.5    & 0.63    \\
            TokenSkip ($\gamma=0.5$)    & 75.2     & 72.4      & 70.1    & 48.5       & 44.6        & 30.9    & 0.55    \\
            \textbf{Adaptive GoGI-Skip} & 93.1     & 91.3      & 88.2    & 60.5       & 55.0        & 23.7    & 0.52    \\
            \bottomrule
        \end{tabular}
    } 
\end{table*}

To elucidate pruning behavior, we analyzed average token retention rates across functional token categories for 7,472 validated MATH training samples (Gemma3-4B-Instruct). We categorized tokens by linguistic and mathematical roles.

Table \ref{tab:appendix_retention_detail} reports the retention rates. Adaptive GoGI-Skip exhibits a discriminative profile: it achieves the highest retention for Numerals (93.1\%), Operators (91.3\%), and Symbols (88.2\%), validating that GoGI identifies and preserves tokens essential for mathematical logic. This retention contrasts with TokenSkip ($\gamma=0.5$), which retains only 75.2\% of Numerals.

Conversely, Adaptive GoGI-Skip aggressively removes General Language tokens (23.7\% retention)—the category comprising the bulk of compressible text. This rate is significantly lower than other methods, indicating targeted pruning of descriptive text with low functional relevance. For intermediate categories like Formatting (60.5\%) and Connectives (55.0\%), Adaptive GoGI-Skip maintains a balanced approach. While many tokens in these categories are pruned, a subset critical for coherence (likely preserved by ANC) is retained more effectively than by aggressive static methods like TokenSkip ($\gamma=0.5$), which reduces Formatting and Connectives retention to 48.5\% and 44.6\%, respectively.

Other baselines show distinct profiles. C3oT retains considerably more General Language (48.6\%) and Formatting (63.4\%) tokens than Adaptive GoGI-Skip, suggesting less targeted pruning. Spiritft's retention profile falls between TokenSkip's conservative ($\gamma=0.6$) and aggressive ($\gamma=0.5$) settings. Prompting retains most tokens, reflecting inefficiency.

This analysis demonstrates that Adaptive GoGI-Skip achieves functionally aware compression. High mathematical token retention supports GoGI effectiveness, while balanced structural retention implies ANC success, contributing to the method's superior efficiency-accuracy balance.
\section{Detailed Analysis of Goal-Gradient Importance Scores}
\label{app:gogi_analysis}

This appendix analyzes GoGI scores, extending Section~\ref{ssec:analysis}. We investigated relationships between GoGI scores and external factors (CoT length, problem difficulty) and internal model metrics (predictive entropy). All analyses utilized GoGI scores computed at the selected target layer $l^*$ \S \ref{app:layer_contribution}) for the 7,472 validated MATH training samples (Gemma3-4B-Instruct).

\subsection{Relationship Between GoGI and External Problem Characteristics}
\label{app:gogi_vs_external}

\begin{figure*}[t]
    \centering
    \begin{minipage}{0.98\linewidth}
        \centering
        \includegraphics[width=\linewidth]{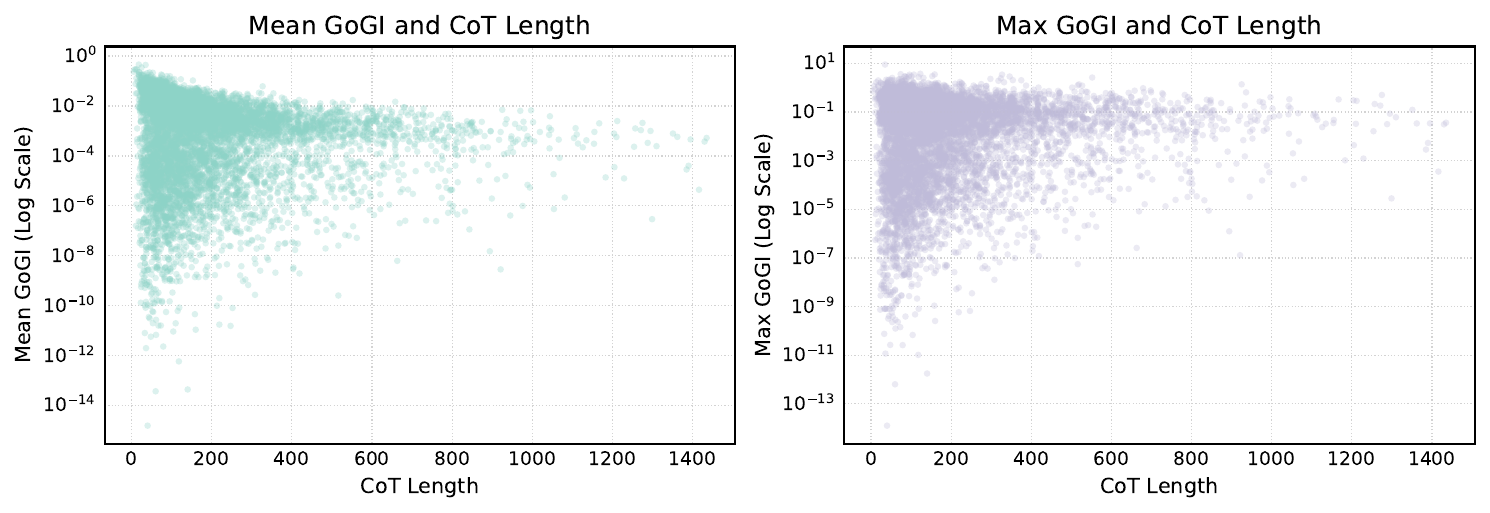}
    \end{minipage}
    \begin{minipage}{0.98\linewidth}
        \centering
        \includegraphics[width=\linewidth]{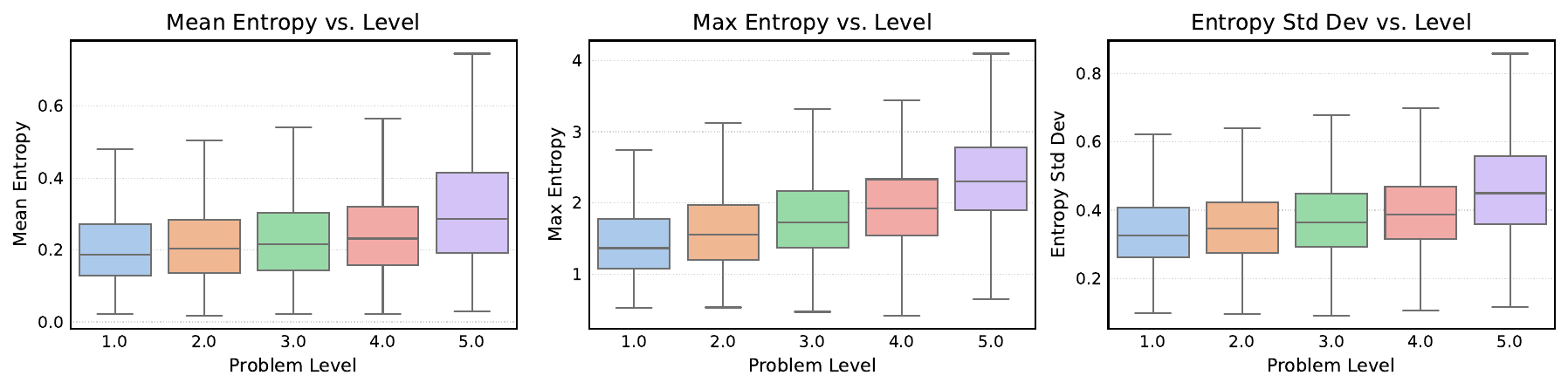}
    \end{minipage}
    \caption{\textbf{GoGI scores vs. external characteristics.} \textbf{Top:} Mean/Max GoGI vs. CoT length. \textbf{Bottom:} Log(1 + Mean GoGI) vs. difficulty level. Weak correlations suggest GoGI captures functional importance independent of these factors.}
    \label{fig:appendix_gogi_vs_external}
\end{figure*}

To determine if GoGI scores reflect superficial sequence properties, we investigated their relationship with CoT length and problem difficulty (Figure \ref{fig:appendix_gogi_vs_external}).

Scatter plot analysis (Figure \ref{fig:appendix_gogi_vs_external}, top) shows Mean GoGI scores exhibit only a weak negative correlation with CoT length (Spearman's $r_s \approx -0.15$, $p < 0.01$), implying minimal impact from sequence length on average token importance. Crucially, Maximum GoGI scores show negligible correlation with sequence length ($|r_s| < 0.05$, n.s.), indicating that highly critical token occurrence is independent of reasoning chain length.

Regarding difficulty (Figure \ref{fig:appendix_gogi_vs_external}, bottom), we found no significant monotonic relationship between GoGI metrics and MATH problem difficulty. Although harder problems showed slightly increased GoGI variance, median importance values remained stable across difficulty strata.

These findings indicate that GoGI captures functional importance transcending superficial characteristics. This evidence supports dynamic strategies like ADS over static heuristics based on length or difficulty.

\subsection{Statistical Independence from Model Metrics}
\label{app:gogi_correlations}

\begin{figure}[t]
    \centering
    \includegraphics[width=\linewidth]{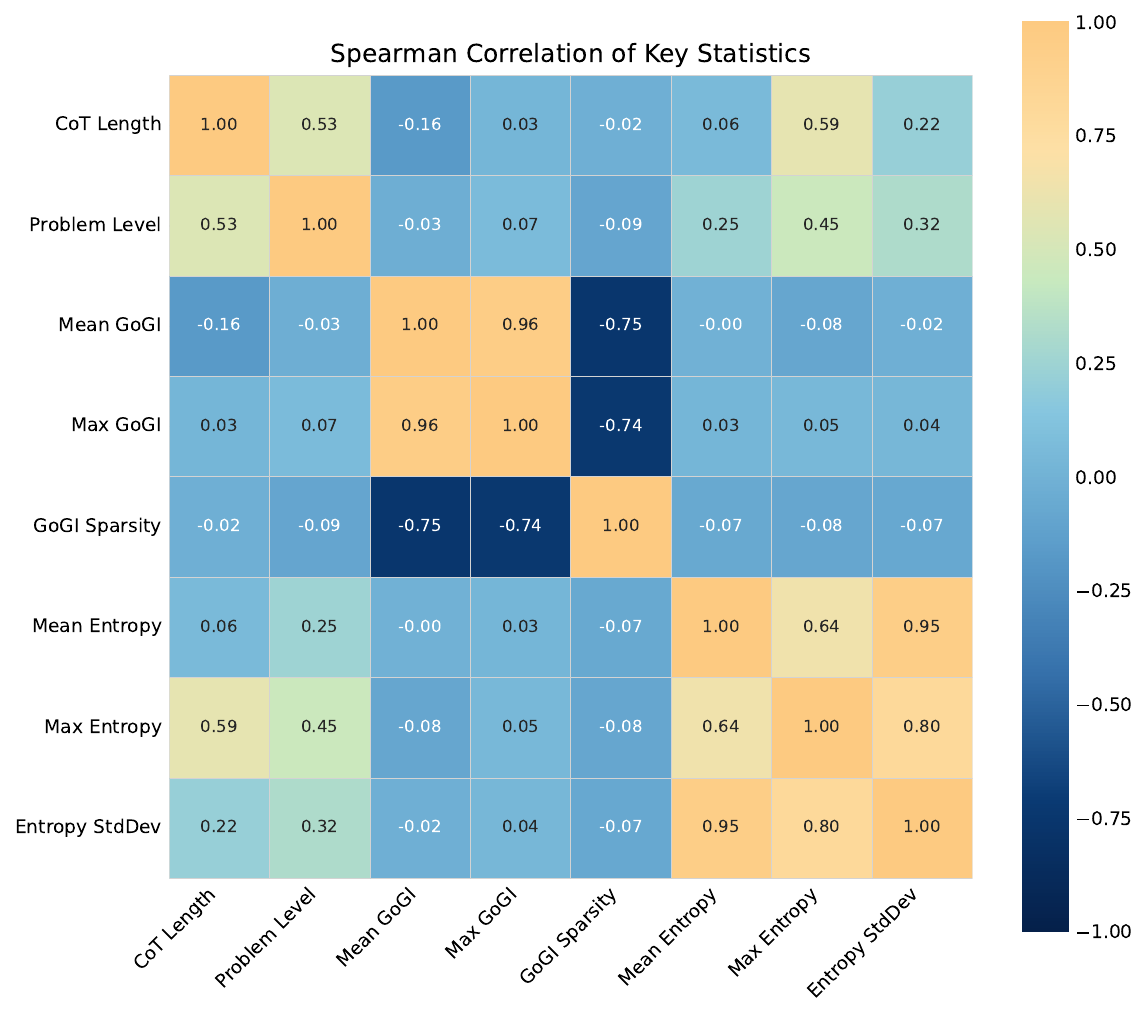}
    \caption{Spearman rank correlation matrix quantifying relationships among statistical measures derived from CoT samples.}
    \label{fig:correlation_matrix}
\end{figure}

To quantify GoGI's unique information relative to predictive entropy, we computed the Spearman rank correlation matrix across GoGI- and entropy-based statistics (Figure \ref{fig:correlation_matrix}).

The analysis confirms key hypotheses:

\paragraph{Internal Consistency.}
GoGI measures exhibit high positive intercorrelations ($r_s > 0.8$), indicating robust internal consistency.

\paragraph{Sparsity Relationship.}
GoGI sparsity shows a strong negative correlation with other GoGI metrics, confirming functional importance concentration in few tokens.

\paragraph{Orthogonality to Uncertainty.}
Critically, correlations between core GoGI metrics and entropy-based uncertainty measures (mean, max, std. dev.) consistently show minimal association ($|r_s|$ typically $< 0.1$), providing evidence for their statistical independence.

This near-orthogonality ($|r_s| < 0.1$) justifies synergistically leveraging these distinct signals within Adaptive Dynamic Skipping (ADS). This confirms the weak associations between GoGI and external characteristics noted in \S \ref{app:gogi_vs_external}.

\begin{figure}[t]
    \centering
    \includegraphics[width=\linewidth]{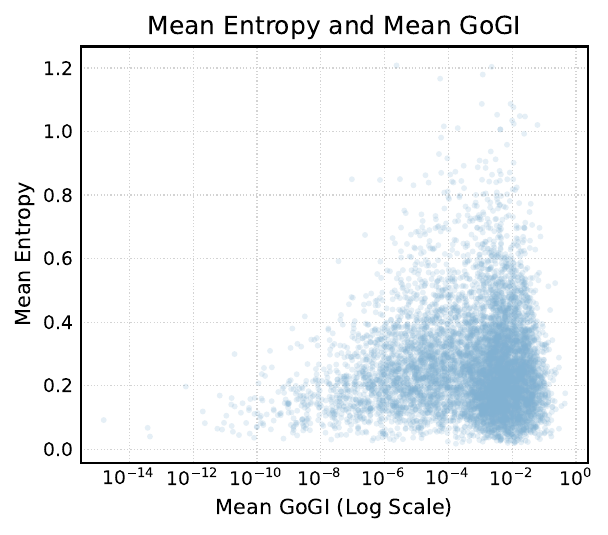}
    \caption{Mean Entropy vs. Mean GoGI score per CoT sample. The lack of trend supports statistical independence.}
    \label{fig:entropy_vs_mean_gogi}
\end{figure}

\subsection{Sample-Level Relationship between GoGI and Entropy}
\label{app:gogi_entropy_sample}

We also examined the relationship visually. While Figure \ref{fig:gogi_and_dynamic} indicates weak per-token correlation, Figure \ref{fig:entropy_vs_mean_gogi} confirms this independence at the sequence level.

The scatter plot reveals no systematic relationship between average functional importance and average predictive uncertainty. Sequences with high average importance can exhibit high or low uncertainty. This reinforces that GoGI and entropy capture fundamentally different reasoning aspects, justifying their combined use in ADS.
\section{Detailed Entropy Analysis}
\label{app:entropy_analysis}

\begin{figure*}[t]
    \centering
    \begin{minipage}{0.98\linewidth}
        \centering
        \includegraphics[width=\linewidth]{figure/entropy_distributions.pdf}
    \end{minipage}
    \begin{minipage}{0.98\linewidth}
        \centering
        \includegraphics[width=\linewidth]{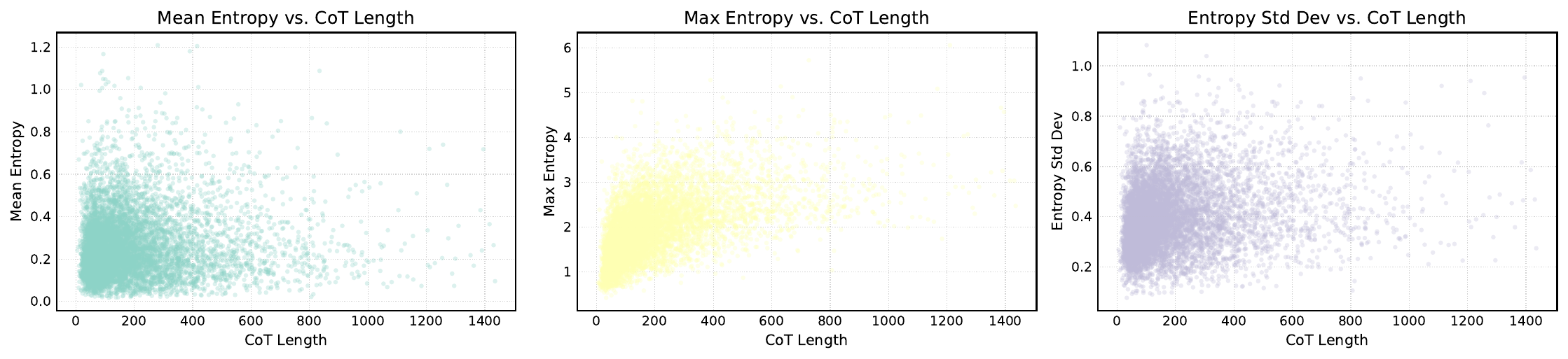}
    \end{minipage}
    \begin{minipage}{0.98\linewidth}
        \centering
        \includegraphics[width=\linewidth]{figure/entropy_vs_level.pdf}
    \end{minipage}
    \caption{\textbf{Statistical analysis of predictive entropy ($H_t$) on MATH.} \textbf{Top:} Empirical distributions of Mean, Max, and Std. Dev. Entropy, showing right-skewed patterns. \textbf{Middle:} Scatter plots showing minimal linear correlation between entropy metrics and CoT sequence length. \textbf{Bottom:} Box plots demonstrating a consistent positive monotonic association between entropy metrics and problem difficulty.}
    \label{fig:appendix_entropy_analysis}
\end{figure*}

This appendix presents a statistical analysis of predictive entropy ($H_t$) across CoT sequences from the MATH training dataset (Gemma3-4B-Instruct). Characterizing $H_t$ is crucial because it functions as the primary runtime signal for uncertainty quantification within our ADS mechanism, informing both EDR and ANC components. Figure \ref{fig:appendix_entropy_analysis} illustrates key distributional properties and relationships with external problem attributes.

\subsection{Distributional Characteristics}
\label{app:entropy_distribution}

The empirical distributions for entropy statistics (Figure \ref{fig:appendix_entropy_analysis}, top) reveal distinct patterns. Mean Entropy exhibits a right-skewed distribution; the majority of reasoning sequences demonstrate relatively low average uncertainty, yet the distribution extends tails toward higher uncertainty regions. Maximum Entropy similarly displays right skewness but spans a considerably wider range, confirming the occurrence of significant transient uncertainty peaks within individual sequences.

The distribution of Entropy Standard Deviation reveals typical intra-sequence variability and provides scaling information for the mapping functions $\mathcal{M}$ and $\mathcal{M}'$. These characteristics inform the statistical normalization within the ADS mechanism (Sections \ref{sssec:edr} and \ref{sssec:anc}). Specifically, we use the empirically estimated global parameters $H_{\text{median}}$ and $H_{\text{std}}$ to standardize entropy inputs.

\subsection{Relationship with External Factors}
\label{app:entropy_relationships}

We examined the relationship between entropy statistics and CoT sequence length via regression analysis (Figure \ref{fig:appendix_entropy_analysis}, middle). We observed consistently weak linear correlations across all statistics. These minimal correlations demonstrate that sequence length predicts model uncertainty poorly, reinforcing our argument that length-based heuristics cannot adapt to variable reasoning complexity.

In contrast, the relationship between entropy and problem complexity (Figure \ref{fig:appendix_entropy_analysis}, bottom) differs markedly. We observe a statistically significant positive monotonic relationship between entropy metrics and MATH problem difficulty. Mean, Maximum, and Standard Deviation of Entropy all systematically increase with problem difficulty. This indicates that as problems become objectively more complex, the model experiences higher average uncertainty, more pronounced peaks, and greater variability.

This correlation between predictive entropy and intrinsic problem complexity validates using predictive entropy to modulate compression intensity. Unlike static heuristics based on superficial characteristics, predictive entropy effectively captures the underlying cognitive demands of varying task complexity, enabling principled adaptation.
\section{Adaptive N-Constraint Mechanism Details}
\label{app:anc_details}

\begin{table*}[t]
    \centering
    \caption{\textbf{Quantitative Analysis of Token Preservation by ANC.} Analysis of ~405k tokens from the MATH test set (Gemma3-4B-Instruct). \textit{Initially Pruned}: tokens where $\mathcal{G}'_t < \tau_t$. \textit{Preserved by ANC}: theoretically pruned tokens retained by N-constraint. \textit{Preservation Rate}: percentage of initially pruned tokens preserved.}
    \label{tab:anc_retention_stats}
    \resizebox{\textwidth}{!}{
        \begin{tabular}{lcccc}
            \toprule
            \textbf{Functional Category}  & \textbf{Total Count} & \textbf{Initially Pruned} & \textbf{Preserved by ANC} & \textbf{Preservation Rate} \\
            \midrule
            Mathematical Numerals         & 50k                  & 5.3k                      & 1.5k                      & 30\%                       \\
            Mathematical Operators        & 40k                  & 5.7k                      & 2.1k                      & 33\%                       \\
            Mathematical Symbols          & 30k                  & 4.5k                      & 1.3k                      & 29\%                       \\
            Structural Formatting         & 60k                  & 24.0k                     & 9.6k                      & 40\%                       \\
            Logical Connectives           & 25k                  & 11.0k                     & 4.8k                      & 44\%                       \\
            General Language Tokens       & 200k                 & 130.0k                    & 39.0k                     & 30\%                       \\
            \midrule
            \textbf{Aggregate Statistics} & 405k                 & 180.5k                    & 58.3k                     & 32\%                       \\
            \bottomrule
        \end{tabular}
    }
\end{table*}

This appendix specifics the ANC mechanism, a critical component of Adaptive GoGI-Skip preserving local coherence. We present a quantitative analysis of token preservation patterns and detail algorithm refinements.

\subsection{ANC Trigger Analysis by Token Type}
\label{app:anc_trigger_analysis}

Table \ref{tab:anc_retention_stats} summarizes token preservation patterns when ANC triggers. Although General Language Tokens account for the highest absolute preservation count, the highest rates occur for Logical Connectives (44\%) and Structural Formatting (40\%). This evidence supports our hypothesis that ANC preferentially safeguards tokens essential for structural integrity and logical coherence, even when they possess low GoGI scores.

The substantial preservation rates for mathematical elements (Numerals, Operators, Symbols, averaging $\sim$30-33\%) further demonstrate ANC effectiveness in preventing fragmentation of essential calculation sequences, particularly when intermediate tokens exhibit low isolated GoGI scores. This confirms ANC acts as a crucial coherence-preserving complement to goal-oriented filtering mechanisms (GoGI, EDR).

\subsection{Algorithmic Refinements}
\label{app:anc_refinements}

The core Adaptive GoGI-Skip algorithm (Section \ref{ssec:pruning_algorithm}) integrates GoGI, EDR, and ANC. We implemented several refinements to enhance robustness and responsiveness:

\paragraph{Entropy Gradient Detection.}
We monitor the change in normalized entropy, $\Delta \hat{H}_t = \hat{H}_t - \hat{H}_{t-1}$. If $|\Delta \hat{H}_t| > \delta_{high}$ ($\delta_{high}=0.3$), indicating a sharp uncertainty transition, we reduce $N_t$ (multiplying by 0.8, clamped to $N_{\min}$) to enforce stricter coherence. Conversely, if $|\Delta \hat{H}_t| < \delta_{low}$ ($\delta_{low}=0.05$) for 3 consecutive steps (indicating stability), we increase $N_t$ (clamped to $N_{\max}$) to allow aggressive pruning.

\paragraph{Compression Rate Modulation of N.}
We modulate $N_t$ based on the global target compression ratio $\gamma_{target}$. Aggressive compression (lower $\gamma_{target}$) increases the effective $N_t$ (scaling $s_N > 1$), while conservative targets decrease it. This aligns local coherence constraints with global efficiency objectives.

\paragraph{Extreme Low-Importance Override.}
To prevent preserving genuinely irrelevant tokens, we prune any token with GoGI score $\mathcal{G}'_t < \theta_{critical} \cdot \tau_t$ ($\theta_{critical}=0.4$), irrespective of consecutive skip count $C_{t-1}$ or current N-constraint $N_t$.

\paragraph{Boundary Condition Management.}
We employ edge-value padding for robust entropy estimation near sequence boundaries when calculating $\bar{H}_{t,W}$.

These refinements collectively improved dynamic pruning stability and effectiveness in preliminary experiments; all reported results utilize this refined algorithm.
\section{Quantitative Validation of Dynamic Compression Mechanisms}
\label{app:dynamic_mech_validation}

This appendix provides a quantitative analysis of the multi-factor dynamic compression behavior produced by our ADS mechanism, based on Figure \ref{fig:3d_analysis}. This visualization plots the effective token retention rate ($1 - \text{compression ratio}$) against normalized runtime entropy and normalized CoT length, stratified by global GoGI score distribution profiles.

\subsection{Analysis of Multi-dimensional Compression Dynamics}

Figure \ref{fig:3d_analysis} reveals the interplay between GoGI context, entropy, and sequence length in governing ADS pruning decisions.

\paragraph{Entropy-Driven Regulation.}
The visualization demonstrates a positive association between normalized entropy (uncertainty) and token retention rate ($\gamma_t$). As entropy increases (vertical axis), the surface plot exhibits consistently higher retention (brighter color/higher elevation), indicating conservative pruning under heightened uncertainty. This confirms the EDR principle: the framework adaptively preserves content when facing uncertainty. The non-linear response reflects the specific mapping function $\mathcal{M}$ (Appendix \ref{app:setup_details}).

\paragraph{Length-Sensitivity Analysis.}
The influence of normalized CoT length is nuanced. While increasing length occasionally correlates with decreased retention (suggesting longer sequences offer compression opportunities), this trend is non-uniform and varies considerably across entropy and importance regimes.

\paragraph{GoGI Distribution Effects.}
Comparing panels illustrates the foundational influence of importance distribution. Higher GoGI concentration consistently yields higher baseline retention across the entropy/length space. This confirms that the global GoGI profile establishes the importance context, which local predictive entropy modulates via EDR.

\subsection{Integrated Interpretation}

This analysis empirically validates that the compression strategy emerges from the synergistic integration of three signals:
\begin{itemize}
    \item Token-specific functional importance (GoGI score $\mathcal{G}'_t$).
    \item Real-time model uncertainty (predictive entropy $H_t$).
    \item Broader sequence context (captured by factors like normalized length).
\end{itemize}

This multi-dimensional framework enables nuanced compression decisions that balance efficiency against the preservation of critical information. The parameter interactions visually substantiate our theoretical model, providing evidence supporting dynamic approaches over static strategies incapable of accommodating context.
\section{Qualitative Case Studies of CoT Compression Methods}
\label{app:case_studies}

This appendix presents a qualitative analysis comparing CoT compression strategies. We use a representative MATH dataset example to elucidate the operational characteristics of Adaptive GoGI-Skip relative to baseline methods, complementing the quantitative evaluations in Section \ref{sec:experiment}.

\subsection{Case Study: MATH Training Sample}

\begin{center}
    \begin{tcolorbox}[
            title=Problem,
            colbacktitle=mytitleindigo,
            coltitle=white,
            colframe=mytitleindigo,
            colback=white,
            arc=2mm,
            boxrule=0.7pt,
            left=1mm,
            right=1mm,
            top=1mm,
            bottom=1mm
        ]
        \begin{lstlisting}
The sequence of integers in the row of squares and in each of the two columns of squares form three distinct arithmetic sequences. What is the value of $N$?

[asy]
unitsize(0.35inch);
ndraw((0,0)--(7,0)--(7,1)--(0,1)--cycle);
ndraw((1,0)--(1,1));
ndraw((2,0)--(2,1));
ndraw((3,0)--(3,1));
ndraw((4,0)--(4,1));
ndraw((5,0)--(5,1));
ndraw((6,0)--(6,1));
ndraw((6,2)--(7,2)--(7,-4)--(6,-4)--cycle);
ndraw((6,-1)--(7,-1));
ndraw((6,-2)--(7,-2));
ndraw((6,-3)--(7,-3));
ndraw((3,0)--(4,0)--(4,-3)--(3,-3)--cycle);
ndraw((3,-1)--(4,-1));
ndraw((3,-2)--(4,-2));
label("21",(0.5,0.8),S);
label("14",(3.5,-1.2),S);
label("18",(3.5,-2.2),S);
label("$N$",(6.5,1.8),S);
label("-17",(6.5,-3.2),S);
[/asy]
\end{lstlisting}
    \end{tcolorbox}

    \begin{tcolorbox}[
            title=Original CoT (count: 156),
            colbacktitle=mytitlered,
            coltitle=white,
            colframe=mytitlered,
            colback=white,
            arc=2mm,
            boxrule=0.7pt,
            left=1mm,
            right=1mm,
            top=1mm,
            bottom=1mm
        ]
        \begin{lstlisting}
Since $18 - 14 = 4$, the common difference in the first column of squares is 4, so the number above 14 is $14 - 4 = 10$, and the number above 10 is $10 - 4 = 6$. This is also the fourth number in the row, so the common difference in the row is $(6 - 21)/3 = -5$.

Then the seventh (and last) number in the row is $21 - 5 \cdot 6 = -9$. In the second column, the common difference is [(-17) - (-9)]/4 = -2, so $N = -9 - (-2) =
\end{lstlisting}
    \end{tcolorbox}
    \captionof{figure}{Problem statement and original Chain-of-Thought (CoT).}
    \label{fig:case_study_p1}
\end{center}

\begin{center}
    \begin{tcolorbox}[
            title=Prompting Baseline (count: 118),
            colbacktitle=mytitleorange,
            coltitle=white,
            colframe=mytitleorange,
            colback=white,
            arc=2mm,
            boxrule=0.7pt,
            left=1mm,
            right=1mm,
            top=1mm,
            bottom=1mm
        ]
        \begin{lstlisting}
The common difference for the first column is 4 (18-14). The number above 14 is 10 (14-4), and above that is 6 (10-4). This is the 4th number in the row. The common difference for the row is -5 ((6-21)/3). The 7th number is -9 (21 - 5*6). The difference for the second column is -2 ((-17 - -9)/4). Therefore, N = -9 - (-2) =
\end{lstlisting}
    \end{tcolorbox}

    \begin{tcolorbox}[
            title=C3oT Baseline (count: 106),
            colbacktitle=mytitleyellow,
            coltitle=white,
            colframe=mytitleyellow,
            colback=white,
            arc=2mm,
            boxrule=0.7pt,
            left=1mm,
            right=1mm,
            top=1mm,
            bottom=1mm
        ]
        \begin{lstlisting}
The common difference in the first column is $18 - 14 = 4$. The number above 14 is $14 - 4 = 10$, above 10 is $10 - 4 = 6$. This is the fourth number in the row. The common difference in the row is $(6 - 21)/3 = -5$. The seventh number is $21 - 5 \cdot 6 = -9$. In the second column, the difference is $[(-17) - (-9)]/4 = -2$. So $N = -9 - (-2) =
\end{lstlisting}
    \end{tcolorbox}

    \begin{tcolorbox}[
            title=Spiritft Baseline (count: 109),
            colbacktitle=mytitlegreen,
            coltitle=white,
            colframe=mytitlegreen,
            colback=white,
            arc=2mm,
            boxrule=0.7pt,
            left=1mm,
            right=1mm,
            top=1mm,
            bottom=1mm
        ]
        \begin{lstlisting}
Since $18 - 14 = 4$, common difference first column 4, number above 14 is $14 - 4 = 10$, number above 10 is $10 - 4 = 6$. This is fourth number in row, common difference row is $(6 - 21)/3 = -5$. Then seventh last number row $21 - 5 \cdot 6 = -9$. In second column, common difference $[(-17) - (-9)]/4 = -2$, so $N = -9 - (-2) =
\end{lstlisting}
    \end{tcolorbox}

    \begin{tcolorbox}[
            title=TokenSkip ~0.5 Baseline (count: 105),
            colbacktitle=mytitleblue,
            coltitle=white,
            colframe=mytitleblue,
            colback=white,
            arc=2mm,
            boxrule=0.7pt,
            left=1mm,
            right=1mm,
            top=1mm,
            bottom=1mm
        ]
        \begin{lstlisting}
$18 - 14 = 4$ common difference first column 4 above 14 $14 - 4 = 10$ above 10 $10 - 4 = 6$ fourth number row common difference $(6 - 21)/3 = -5$ seventh last $21 - 5 6 = -9$ second column difference $[(-17) - (-9)]/4 = -2$ $N = -9 - (-2) =
\end{lstlisting}
    \end{tcolorbox}

    \begin{tcolorbox}[
            title=\textbf{Adaptive GoGI-Skip (count: 87)},
            colbacktitle=mytitleviolet,
            coltitle=white,
            colframe=mytitleviolet,
            colback=white,
            arc=2mm,
            boxrule=0.7pt,
            left=1mm,
            right=1mm,
            top=1mm,
            bottom=1mm
        ]
        \begin{lstlisting}
18 -14 = 4 in squares 4 so 14 $14 -4 =10 above $10 -6$. fourth so row $(6 -21)/3 = -5$. seventh and the is $21 -5 \cdot 6 = -9$. the is $[(-17) - (-9)]/4 = -2$, so $N = -9 - (-2) =
\end{lstlisting}
    \end{tcolorbox}
    \captionof{figure}{Comparison of compressed CoT sequences by method.}
    \label{fig:case_study_p2}
\end{center}

\subsection{Comparative Analysis of Compression Strategies}

This analysis highlights distinct operational characteristics of the evaluated approaches:

\paragraph{Original CoT (156 tokens).}
Provides a complete, syntactically coherent explanation featuring explicit verbal connectors (``Since'', ``so'', ``Then'', ``In'') and thorough elaborations of mathematical reasoning principles.

\paragraph{Prompting Baseline (118 tokens).}
Retains most semantic content but uses concise phrasing and parenthetical notation (e.g., `(18-14)`). This achieves moderate compression (24\% reduction) with minor structural modification.

\paragraph{C3oT Baseline (106 tokens).}
Systematically removes select function words (``is'') and redundant phrases while preserving core mathematical notation and sentence structure. This results in 32\% reduction with minimal grammatical disruption.

\paragraph{SpiritFT Baseline (109 tokens).}
Applies aggressive linguistic element reduction (removing ``Since'', articles, auxiliary verbs) while maintaining critical mathematical expressions and basic sentence structure, achieving 30\% reduction.

\paragraph{TokenSkip ($\gamma \approx 0.5$, 105 tokens).}
Reduces explanatory language significantly, yielding keyword/numerical sequences. It prioritizes reserving mathematical operators and numerical relationships (33\% reduction).

\paragraph{Adaptive GoGI-Skip (87 tokens).}
Achieves maximal efficiency (44\% reduction) via precise identification of functionally critical tokens. It systematically preserves essential numerical expressions ($18-14=4$, $14-4=10$, $(6-21)/3=-5$) while aggressively pruning descriptive language, connectives, and non-contributory spacing.

This analysis highlights how Adaptive GoGI-Skip's goal-oriented importance scoring facilitates effective identification of mathematically essential tokens, resulting in the most efficient compression among evaluated methods while preserving solution integrity. Unlike static approaches, Adaptive GoGI-Skip's adaptive modulation demonstrates superior context-aware compression capacity.

\end{document}